\documentclass[10pt,journal]{IEEEtran}
\IEEEoverridecommandlockouts

\usepackage{graphicx}
\usepackage{subfigure}
\usepackage{amsmath,amsthm,amsfonts,amssymb}
\usepackage{cite}
\usepackage{bm}
\usepackage{bbm}
\usepackage{url}
\usepackage{array}
\usepackage{color}
\usepackage{multirow}
\usepackage{booktabs}
\usepackage[table,xcdraw]{xcolor}
\usepackage{enumerate}
\usepackage{enumitem}
\usepackage{makecell}

\usepackage[english]{babel}
\usepackage{algorithm}
\usepackage[noend]{algpseudocode}

\theoremstyle{plain}
\newtheorem{thm}{Theorem}
\newtheorem{lem}[thm]{Lemma}

\newtheorem{prop}[thm]{Proposition}

\newtheorem{rem}{Remark}
\newtheorem{sty1}{Theorem}
\newtheorem{defi}[sty1]{Definition}

\newenvironment{NewProof}{{\noindent\it Proof.}}{\hfill $\blacksquare$\par}
\newcommand{\RR}{I\!\!R}

\usepackage[english]{babel}
\usepackage{algorithm}
\usepackage[noend]{algpseudocode}

\allowdisplaybreaks[4]

\begin{document}
\title{Denoising Noisy Neural Networks: A Bayesian Approach with Compensation}

\author{
Yulin~Shao,~\IEEEmembership{Member,~IEEE},
Soung Chang Liew,~\IEEEmembership{Fellow,~IEEE},
Deniz~G\"und\"uz,~\IEEEmembership{Fellow,~IEEE}
\thanks{Y. Shao and D. G\"und\"uz are with the Department of Electrical and Electronic Engineering, Imperial College London, London SW7 2AZ, U.K. (e-mail: y.shao@imperial.ac.uk, d.gunduz@imperial.ac.uk).
}
\thanks{S. C. Liew is with the Department of Information Engineering, The Chinese University of Hong Kong, Shatin, New Territories, Hong Kong (e-mail: soung@ie.cuhk.edu.hk).
}
\thanks{This work was supported in part by the European Research Council project BEACON under grant number 677854, and in part by CHIST-ERA grant CHIST-ERA-18-SDCDN-001 (funded by EPSRC-EP/T023600/1).
This work was also supported by a CUHK Direct Grant (Project No: 4055157).}
}

\maketitle

\begin{abstract}
Deep neural networks (DNNs) with noisy weights, which we refer to as noisy neural networks (NoisyNNs), arise from the training and inference of DNNs in the presence of noise.
NoisyNNs emerge in many new applications, including the wireless transmission of DNNs, the efficient deployment or storage of DNNs in analog devices, and the truncation or quantization of DNN weights.
This paper studies a fundamental problem of NoisyNNs:
how to reconstruct the DNN weights from their noisy manifestations.
While prior works relied exclusively on the maximum likelihood (ML) estimation, this paper puts forth a denoising approach to reconstruct DNNs with the aim of maximizing the inference accuracy of the reconstructed models.
The superiority of our denoiser is rigorously proven in two small-scale problems, wherein we consider a quadratic neural network function and a shallow feedforward neural network, respectively.
When applied to advanced learning tasks with modern DNN architectures, our denoiser exhibits significantly better performance than the ML estimator.
Consider the average test accuracy of the denoised DNN model versus the weight variance to noise power ratio (WNR) performance.
When denoising a noisy ResNet34 model arising from noisy inference, our denoiser outperforms ML estimation by up to $4.1$ dB to achieve a test accuracy of $60\%$.
When denoising a noisy ResNet18 model arising from noisy training, our denoiser outperforms ML estimation by $13.4$ dB and $8.3$ dB to achieve test accuracies of $60\%$ and $80\%$, respectively.
\end{abstract}


\section{Introduction}\label{sec:intro}
The past decade has witnessed a fundamental paradigm shift in a variety of disciplines driven by deep learning (DL) technologies \cite{DL}.
Unlike traditional techniques that use human-crafted models, DL leverages deep neural networks (DNNs), which are parametric models with a huge number of weights/parameters, to learn features and patterns from data. 
Although DL-based solutions have achieved remarkable success in a wide range of applications \cite{he2016deep,Transformer,AlphaGo,Significant2020}, they also face new challenges when applied in practice. The training and inference of DNNs in the presence of noise, which we refer to as noisy neural networks (NoisyNNs), is one of the challenges that call for urgent attention.

\subsection{Applications}
Noise is inherent in almost all physical media, phenomena, and manipulations. NoisyNNs emerge in applications where the DNNs are transmitted over a wireless channel, deployed on a noisy medium, or when the DNN parameters are truncated or quantized to fit the finite system precision.

{\it \textbf{NoisyNN incurred by wireless channels}}:
Real-world DL applications often involve distributed learning and inference among multiple agents. Wireless communication is thus an indispensable part of such learning systems. Wireless channels, however, introduce additive white Gaussian noise (AWGN) to signals passing through it \cite{MLintheair,PNC}.

In federated edge learning (FEEL) \cite{FedAvg,Gunduz:CL:20}, for example, data are generated at the edge devices and a good DNN model can be obtained only when the devices learn collaboratively, with the help of a parameter server (PS). 
Since the training data often contain private information and cannot be shared with the PS or other devices, the information transmitted over the noisy communication channels is often the locally trained DNN models, i.e., the weights that parameterize the DNNs.
Another example is the remote model download. One of the main benefits of DL technologies is their ability to adapt to the physical environment, which is often impossible to model explicitly. This, in turn, requires training of separate DNNs for different physical environments, e.g., for every cell in a cellular network \cite{LIDAR1,LIDAR2}. These well-trained DNNs are maintained at base stations (BSs). A DNN user, e.g., an autonomous vehicle, must download the models wirelessly from the BSs as it enters their coverage areas to adapt to the changing environment.

Due to the increasingly large scale of today's DNNs, prompt and efficient delivery of DNNs over the wireless channel is unachievable with conventional digital communications.
In FEEL, the uplink model aggregation has been well recognized as a bottleneck \cite{FedAvg}. Analog over-the-air computation (OAC) \cite{Gunduz1,Gunduz2,zhu,ShanghaiTech} has been proposed to address this problem.
Also, in terms of remote DNN download, \cite{AirNet} demonstrated that a joint source-and-channel coding (JSCC) approach with analog transmission is more efficient than the conventional digital approaches.
With analog transmission of DNNs, the received DNN is a noisy version of the transmitted one, giving rise to NoisyNN.

{\it \textbf{NoisyNN incurred by analog devices}}:
The deployment of DNNs on analog hardware \cite{Nature1,Nature2,zhou1,electrical,photons} also produces NoisyNNs. In this application, the computations inside a DNN are executed in the analog domain with digital weights being represented by analog quantities, e.g., conductance \cite{Nature1}, electrical voltages \cite{electrical}, or photons \cite{photons}. Compared with digital hardware, analog hardware promises at least two orders of magnitude greater gains in both computational speed and energy efficiency \cite{photons,Nature1,zhou1}. Analog quantities, however, are subject to thermal noise generated by their physical components -- often the deployed weights are different from the expected weights on an analog device. In other words, analog computations are inherently noisy and the deployed neural networks are NoisyNNs. In particular, the noise in analog hardware is often modeled as AWGN to mimic the overall effect of many concurrent random effects. 

Moreover, the efficient storage of DNN weights in analog hardware faces the same problem: the retrieved DNN from an analog device is a noisy version of the stored DNN \cite{Isik}, due to the noisy nature of analog hardware. 

{\it \textbf{NoisyNN incurred by truncation and quantization}}:
In addition to the above analog communication and storage scenarios, in conventional digital systems, the precision loss of NN weights due to quantization \cite{Gray:IT:98} and truncation \cite{Truncate} can also be viewed as a kind of noise. Although this kind of noise is deterministic, it is shown to behave like Gaussian noise in the high-rate compression regime for various types of quantization and compression schemes \cite{Lee:IT:96, Zamir:IT:96, Kipnis:ISIT:19}, implying that the average estimation error achieved when the compressed representation of data is available is asymptotically equivalent to the one achieved from a Gaussian-noise corrupted version.

In general, we can group NoisyNNs into two classes:
\begin{enumerate}
    \item \textbf{NoisyNNs arising from noisy inference}, where the training phase is noiseless, but noise is introduced in the inference phase on the already-trained DNN weights. Typical examples include the wireless transmission and analog deployment of well-trained DNNs.
    \item \textbf{NoisyNNs arising from noisy training}, where noise is introduced over the course of training, but the inference phase is noiseless. A typical example is wireless collaborative learning.
\end{enumerate}

\subsection{Related work}
In the DL community, much of the research effort has been devoted to the noisy training data \cite{Noisydata}. The study of NoisyNNs, in contrast, is relatively few.
To summarize, these few works are mainly focused on two fundamental problems of NoisyNNs:
i) Understanding NoisyNNs: what is the impact of weight noise on the training and inference of DNNs?
ii) Taming NoisyNNs: how to mitigate the effect of noisy weights?

{\it \textbf{Understanding NoisyNNs}}: The analysis of NoisyNNs (and more generally, noisy machines) dates back to the 1950s when Von Neumann began to analyze the effect of noisy gate circuits \cite{von}. Since then, there have been sporadic studies on NoisyNNs, motivated by different applications.

In the application of stochastic circuits, \cite{geo} and a recent work \cite{zhou2} studied noisy binary neural networks (BNNs).
BNNs are feedforward neural networks with binary inputs and sign function as the activation function of each layer.
The outputs of all intermediate layers, including the BNN outputs, are also binary.
Ref. \cite{geo} studied the sensitivity of the BNN to weight errors using a geometric approach. The authors showed that the bit error rate (BER) of the network output is proportional to the amount of noise added to the weights. Ref. \cite{zhou2} considered a noisy BNN where noise is added to the binary outputs of the activation functions in each layer, but not to the weights. In so doing, the connection between two consecutive layers can be viewed as an $n$-input, $m$-output channel, where $n$ and $m$ are the alphabet size of the two layers. The mutual information loss after passing through one layer can then be measured via an information-theoretic approach.  

For conventional feedforward NoisyNNs, \cite{An1996,Tikhonov} considered weight noise in training and studied the impact of noise on the generalization performance of the learned neural networks. In particular, they focused on how much weight noise is changing the loss functions and demonstrated that weight noise has a regularization effect in training.
Ref. \cite{zhou2019toward}, on the other hand, showed that weight noise can be leveraged to  escape from the local minima in DNN training.
In deep reinforcement learning (DRL), the authors of \cite{DRLNoise} deliberately added parametric noise (with learnable mean and variance) in training to enhance the exploration of a policy network.
These works demonstrated that deliberately introduced noisy weights can be beneficial to DNN training, provided that the amount noise can be adjusted judiciously.

{\it \textbf{Taming NoisyNNs}}:
For most applications in practice, the major source of noise is the environment, such as the wireless channel or the carrier medium of DNNs. Unlike artificially-added noise, environmental noise is often unmanageable and detrimental to both DNN training and inference. An important problem is then how to mitigate the effect of noisy weights.

For NoisyNNs arising from noisy inference, the only technique available in the literature to tame NoisyNNs is training with noise injection \cite{zhou1,AirNet,Isik}. Specifically, to cope with noisy weights in the inference phase, the DNN is retrained with artificially-inserted weight noise so that the retrained weights are more robust to random perturbations. This approach, however, suffers from two main limitations:
i) It does not work for NoisyNNs arising from noisy training, where the training phase itself is noisy and the inference phase is noiseless.
ii) It requires the retraining of DNNs, which is undesired or even impossible in most practical scenarios. In wireless communication applications, for example, a BS or an edge server can store a large number of DNN models, which have been trained elsewhere.
This suggests that the BS may not have the source data to retrain a DNN.
For practicality, it would be desirable to have a generic DNN transmission scheme to combat inference noise, but not to train a DNN for each noise-power region.

For NoisyNNs arising from noisy training, on the other hand, the problem of taming weight noise remains open. The noisy weights are left unprocessed in all prior works \cite{zhu,Gunduz1,Gunduz2,ShanghaiTech,Blind,BayesianOAC,Zhu:TWC:21,misalignedOAC}. 
As a consequence, the training can result in a NoisyNN with poor inference performance, and even diverge in the case of large noise power.

\subsection{Contributions}
This paper studies the fundamental problem of taming NoisyNNs and puts forth a new approach, dubbed {\it denoising}, to mitigate the effect of weight noise for NoisyNNs arising from both noisy training and noisy inference. 
The denoising approach processes the NoisyNNs from the observer/receiver's perspective and is effective for generic DNNs -- it does not require any retraining of DNNs to adapt to a specific target noise power, which is much desired in practice.

The idea of denoising originates from the classical statistical inference problem, i.e., how to estimate the uncontaminated weights from their noisy counterparts?
All prior works \cite{zhu,Gunduz1,Gunduz2,ShanghaiTech,Blind,BayesianOAC,Zhu:TWC:21,misalignedOAC,Nature1,Nature2,zhou1,electrical,photons,Isik,TAMU} take the noisy ob\-ser\-vati\-ons/mani\-fes\-ta\-tions of DNNs directly as the estimated DNN weights. 
This is essentially the maximum likelihood (ML) estimation in the language of statistical inference, as the raw observations maximize the likelihood function of the true DNN weights under AWGN.
However, we emphasize that the ultimate goal of taming NoisyNNs is not to estimate the DNN weights with the highest fidelity, but to reconstruct a DNN that would provide the highest inference accuracy for the underlying task.
For this purpose, this paper demonstrates that the ML estimation is in general suboptimal. By exploiting the statistical characteristics of the DNN weights as a kind of prior information, we devise a compensated minimum mean squared error (MMSE) denoiser by introducing a ``population compensator'' and a ``bias compensator'' to the classical MMSE estimation method. We refer to the resulting denoiser as the $\textup{MMSE}_{pb}$ denoiser.

Two main ingredients of the $\textup{MMSE}_{pb}$ denoiser are Bayesian estimation and compensators for Bayesian estimation:
\begin{enumerate}[leftmargin=0.5cm]
\item \textbf{Bayesian estimation}. Unlike ML estimation, we assume the uncontaminated DNN weights are generated from a statistical distribution and perform Bayesian estimation to minimize the MSE or maximize the {\it a posteriori} probability (MAP) of the estimated weights. Given the complex architecture of today's DNNs, acquiring the true statistical distribution is elusive. Thus, we approximate the statistical distribution by a Gaussian prior parameterized by the sample mean and sample variance of the true DNN weights, and postulate that the DNN weights are sampled from the approximated Gaussian in an independent and identically distributed (i.i.d.) manner. On that basis, we devise an MMSE estimator to minimize the MSE between the estimated weights and the true weights (note that the MMSE estimator is also a MAP estimator since both criteria are equivalent under the assumption of a Gaussian prior).

\item \textbf{Compensators for Bayesian estimation}. An MMSE estimator minimizes the MSE for the estimated DNN weights but does not necessarily maximize the inference accuracy of the reconstructed model. 
This is because the DNN weights with larger magnitudes matter more than those with smaller magnitudes \cite{Pruning1,Pruning2} as far as the inference accuracy is concerned, while the MMSE metric treats each DNN weight equally. In this light, we put forth a population compensator and a bias compensator to the MMSE metric and devise the $\textup{MMSE}_{pb}$ denoiser to denoise the NoisyNNs.
\end{enumerate}

As can be seen, the fundamental difference between estimation and denoising is their goal. An estimator (ML, MMSE or MAP) aims to recover the uncontaminated weights as accurately as possible, whereas a denoiser aims to maximize the inference accuracy of the reconstructed model.
The superiority of our $\textup{MMSE}_{pb}$ denoiser over the ML estimator is rigorously proven in two small-scale problems, wherein we consider a quadratic neural network function and a shallow feedforward neural network, respectively.

Extensive experimental results on advanced learning tasks with modern DNN architectures further verify the superior performance of the $\textup{MMSE}_{pb}$ denoiser in both noisy inference and noisy training.
The experiments are performed on a computer vision task (CIFAR-10 \cite{CIFAR10}) and a natural language processing (NLP) task (SST-2 \cite{SST2}), respectively, with various DNN architectures.
\begin{enumerate}[leftmargin=0.55cm]
\item \textbf{Noisy inference}. We consider three well-trained DNN models (ResNet34 \cite{he2016deep}, ResNet18 \cite{he2016deep}, and ShuffleNet V2 \cite{Shufflenet}) for image classification on the CIFAR-10 dataset and a BERT model \cite{bert} for sentiment analysis on the SST-2 dataset. NoisyNNs are generated by adding AWGN noise to the already-trained DNN weights according to a given weight variance to noise power ratio (WNR).
To achieve a $60\%$ test accuracy on the CIFAT-10 dataset, our denoiser outperforms the ML estimator by up to $4.1$ dB, $3$ dB, and $3.8$ dB on the noisy ResNet34, ResNet18, and ShuffleNet V2 model, respectively.
To achieve a $75\%$ test accuracy on the SST-2 dataset with the BERT model, our denoiser is $1.1$ dB better than the ML estimator.
\item \textbf{Noisy training}. We implement a FEEL system with OAC, where AWGN is introduced in the training phase in each FEEL iteration. When our $\textup{MMSE}_{pb}$ denoiser is used at the receiver to reconstruct the aggregated model, remarkable gains over the ML estimator are observed. For the ShuffleNet V2 model, the average test-accuracy gains are up to $1.7$ dB to attain a test accuracy of $60\%$. For the ResNet18 model, the test-accuracy gains is boosted significantly: to achieve a $60\%$ test accuracy, our denoiser is $13.4$ dB better than the ML estimator; to achieve a $80\%$ test accuracy, our denoiser is $8.3$ dB better than the ML estimator.
\end{enumerate}

{\it Notations} --
Throughout the paper, we use boldface lowercase letters to denote column vectors (e.g., $\bm{x}$, $\bm{w}$) and boldface uppercase letters to denote matrices (e.g., $\bm{W}$, $\bm{Z}$); $(*)^\top$ denotes the transpose of a vector or matrix; $|*|$ denotes the cardinality of a vector.
$\RR$ stands for the set of real numbers;
$\bm{e}$ stands for the all-ones vector;
$\bm{I}_{d\times d}$ stands for the $d$-dimensional identity matrix;
$\delta$ stands for the delta function;
$\mathcal{N}$ stands for the real Gaussian distribution;
$\mathcal{U}$ stands for the uniform distribution.

\section{System Model}\label{sec:2}
\subsection{Noisy neural networks}
We consider an artificial neural network $\mathcal{F}$ with input vector $\bm{x}\in\RR^{d_x}$, output vector $\bm{y}\in\RR^{d_y}$, and parameter vector $\bm{w}\in\RR^{d}$:
$\bm{y}=\mathcal{F}(\bm{x}|\bm{w})$. Given a collection of $N$ training examples $\left\{\bm{x}^n,\allowbreak\bm{y}^n:\allowbreak n=1,\allowbreak 2,3,\allowbreak ...,N\right\}$, the goal of learning is to identify the parameter vector $\bm{w}$ that minimizes a specified loss function $\mathcal{L}\left(\left\{\bm{x}^n,\bm{y}^n\right\},\bm{w} \right)$.
The machine learning process can be broken into two phases:
\begin{enumerate}[leftmargin=0.5cm]
\item \textbf{The training phase}. The training phase typically proceeds over multiple epochs. In the $i$-th epoch, the parameter vector $\bm{w}^{(i)}$ is updated via the back-propagation algorithm in the negative gradient direction:
\begin{eqnarray}
\bm{w}^{(i+1)} = \bm{w}^{(i)} -  \zeta^{(i)} \times \nabla_{\bm{w}} \mathcal{L}\left(\left\{\bm{x}^n,\bm{y}^n\right\},\bm{w}^{(i)} \right),
\end{eqnarray}
where $\zeta^{(i)}$ is the learning rate scaling the magnitude of the gradients. As the training progresses, the updated parameters $\bm{w}^{(i)}$ yield smaller and smaller training loss.
\item \textbf{The inference/deployment phase}. When the training phase is complete, the trained parameter vector $\bm{w}^*$ that minimizes the loss function over the training dataset is deployed to make predictions on unseen data examples.
\end{enumerate}
In practical systems, the parameter vector $\bm{w}$ can be exposed to noise in either the training phase, the inference phase, or both. In general, the observed parameter vector $\bm{r}\in\RR^d$ is a noisy version of the true weight vector:
\begin{eqnarray}\label{eq:r}
\bm{r=w+z},
\end{eqnarray}
where the noise term $\bm{z}$ can often be modeled as AWGN. That is, the elements of $\bm{z}$ are sampled from a Gaussian random variable $\mathcal{Z}\sim\mathcal{N}(\mathcal{Z};0,\sigma^2_z)$ in an i.i.d. manner. With noisy weights, the DNN becomes a NoisyNN.

\subsection{Federated edge learning (FEEL)}
\begin{figure}[t]
  \centering
  \includegraphics[width=0.75\linewidth]{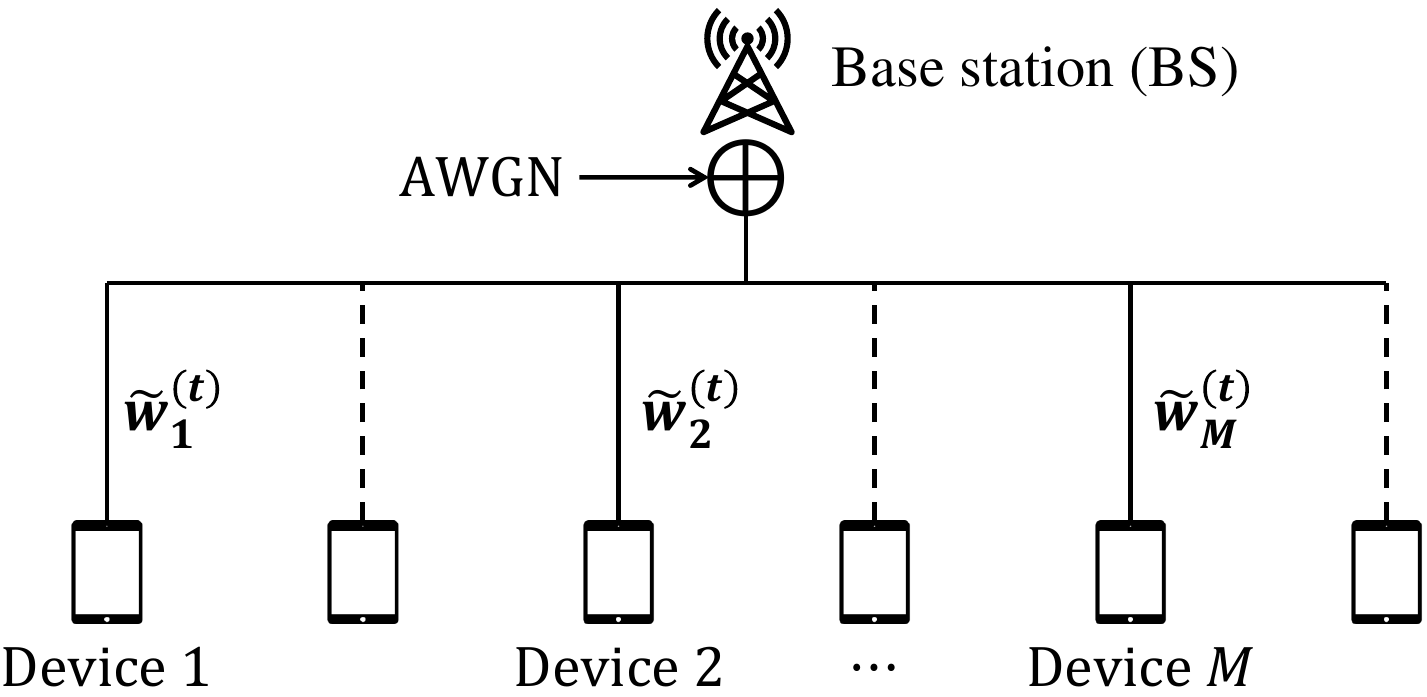}\\
  \caption{In FEEL with OAC, multiple edge devices collaboratively train a common model with the help of a BS (or PS). The uplink model aggregation is realized by OAC.}
\label{fig:1}
\end{figure}
In FEEL, a number of edge devices collaboratively train a shared model with the help of a BS, as shown in Fig.~\ref{fig:1}. Data are distributed at the edge devices and cannot be shared among devices due to privacy concerns. The edge devices train the DNN locally using their local data and transmit the model updates to the BS with over-the-air computation (OAC). One iteration of FEEL operates as follows:
\begin{enumerate}[leftmargin=0.5cm]
\item Downlink broadcast. The BS maintains a global DNN model $\bm{w}^{(t)}\in\RR^d$ and periodically broadcasts the latest model to the edge devices at the beginning of an iteration.
\item Local training. Upon receiving the latest global model, a subset of devices, which are willing to participate in the training in this iteration, train the global model locally using their private dataset for one or more epochs. Let there be $M$ devices participating in the training. Each of the $M$ devices obtains a new model $\bm{w}^{(t)}_m\in\RR^d$, $m=1,2,3,...,M$, after training.
\item Uplink aggregation with OAC. The $M$ devices transmit their model updates $\widetilde{\bm{w}}^{(t)}_m=\bm{w}^{(t)}_m-\bm{w}^{(t)}$ simultaneously to the BS in an analog fashion.\footnote{Analog transmission here does not refer to traditional analog communication techniques such as amplitude or frequency modulation. Instead, it still leverages digital modulation, such as orthogonal frequency division multiplexing (OFDM), but does away with discrete constellations -- the continuous coded symbols are directly mapped onto OFDM subcarriers. This scheme is also known as the discrete-time analog communications in the literature \cite{DiscreteAnalog}.} The signals from different devices overlap at the BS and produce
\begin{eqnarray}\label{eq:OAC}
\bm{r}=\sum_{m=1}^{M}\widetilde{\bm{w}}^{(t)}_m +\bm{z}\triangleq \widetilde{\bm{w}}^{(t)}+\bm{z},
\end{eqnarray}
thanks to the superposition nature of the uplink multiple-access channel (MAC). In particular, $\bm{z}$ is AWGN with a power spectrum density $\sigma^2_z$.
\item {Model update}. Given the received vector $\bm{r}$, the BS estimates the arithmetic sum of the model updates $\widetilde{\bm{w}}^{(t)}=\sum_{m=1}^{M}\widetilde{\bm{w}}^{(t)}_m$ and updates the global model by $\bm{w}^{(t+1)} = \bm{w}^{(t)} + \frac{1}{M} \widetilde{\bm{w}}^{(t)}$.
\end{enumerate}

As can be seen, the edge devices and the BS have to exchange the DNN weights wirelessly over a number of training iterations. In each iteration, AWGN is introduced by the uplink wireless channel and the updated global model $\bm{w}_0$ is a NoisyNN. When the received SNR in \eqref{eq:OAC} is low, the noise can hinder the convergence of FEEL.
\begin{rem}[Fading channels]
To use OAC in fading channels, an additional step before the analog transmission is the channel-coefficient precoding \cite{zhu,Gunduz2,ShanghaiTech}. That is, each device precodes the model updates $\widetilde{\bm{w}}^{(t)}_m$ by the inversion of the uplink channel coefficients to pre-compensate the channel distortion. By doing so, the fading MAC degenerates to a Gaussian MAC, as in \eqref{eq:OAC}.
\end{rem}

\section{Denoising NoisyNNs}\label{sec:3}
In this section, we study how to denoise NoisyNNs and reconstruct the DNN weights from their noisy ob\-ser\-va\-tions/mani\-fes\-ta\-tions $\bm{r}$.

\subsection{ML estimation}
A simple scheme to denoise NoisyNNs is to estimate the DNN weights by the ML estimation, as all prior works did.

\begin{defi}[ML estimation]\label{def:ML}
Given the observed noisy weight vector $\bm{r}\in\RR^d$ in \eqref{eq:r}, the ML estimate of the uncontaminated weight vector $\bm{w}$ is
\begin{equation}\label{eq:ML}
\widehat{\bm{w}}^{\text{ML}} = \bm{r}.
\end{equation}
\end{defi}

The reason behind \eqref{eq:ML} is straightforward.
In ML estimation, the real weight vector $\bm{w}$ is treated as a constant vector.
The likelihood function $p(\bm{r}|\bm{w})$ is then a $d$-dimensional Gaussian $p(\bm{r}|\bm{w})\sim\mathcal{N}(\bm{r};\bm{w},\bm{\Sigma}_z)$, where the covariance matrix $\bm{\Sigma}_z$ is given by $\bm{\Sigma}_z=\sigma^2_z\bm{I}_{d\times d}$, since the elements of $\bm{z}$ are i.i.d. Gaussian noise. 
Therefore, the most likely $\bm{w}$ that has generated the observed weight vector is $\bm{r}$.

In this work, we ask the following question: is there a better approach to reconstruct the DNN weights than the ML estimation? In the following, we answer this question affirmatively by putting forth a Bayesian denoiser with a population compensator and a bias compensator.
To start with, let us consider the Bayesian estimation of $\bm{w}$.

\subsection{Bayesian estimation}
The ML estimation treats $\bm{w}$ as a constant vector and aims to find the ML estimate $\widehat{\bm{w}}^{\text{ML}}$. Bayesian estimation, on the other hand, treats $\bm{w}$ as a random vector sampled from a statistical distribution $p(\bm{w})$ and aims to find either the MMSE estimate $\widehat{\bm{w}}^{\text{MMSE}}$ or the MAP estimate $\widehat{\bm{w}}^{\text{MAP}}$. Specifically, the MMSE estimate $\widehat{\bm{w}}^{\text{MMSE}}$ minimizes the MSE:
\begin{eqnarray}\label{eq:MMSE1}
\widehat{\bm{w}}^{\text{MMSE}}
\hspace{-0.2cm}&=&\hspace{-0.2cm} \arg\min_{\widehat{\bm{w}}} \mathbb{E}\left[\left(\widehat{\bm{w}}-\bm{w}\right)^2|\bm{r} \right] \nonumber\\
\hspace{-0.2cm}&=&\hspace{-0.2cm} \arg\min_{\widehat{\bm{w}}} \int \left(\widehat{\bm{w}}-\bm{w}\right)^2 p(\bm{w},\bm{r}) d\bm{w} \nonumber\\
\hspace{-0.3cm}&=&\hspace{-0.3cm} \arg\min_{\widehat{\bm{w}}} \int \left(\widehat{\bm{w}}-\bm{w}\right)^2 p(\bm{r}|\bm{w})p(\bm{w}) d\bm{w},
\end{eqnarray}
while the MAP estimate $\widehat{\bm{w}}^{\text{MAP}}$ maximizes the posterior probability:
\begin{equation}\label{eq:MAP1}
\widehat{\bm{w}}^{\text{MAP}}
= \arg\max_{\bm{w}} p(\bm{w}|\bm{r})
= \arg\max_{\bm{w}}p(\bm{r}|\bm{w})p(\bm{w}).
\end{equation}

The likelihood function $p(\bm{r}|\bm{w})$ in \eqref{eq:MMSE1} and \eqref{eq:MAP1} is determined by \eqref{eq:r}. Thus, we have $p(\bm{r}|\bm{w})\sim\mathcal{N}(\bm{r};\bm{w},\bm{\Sigma}_z)$, as in the ML estimation. The joint distribution of DNN weights $p(\bm{w})$, on the other hand, describes the interrelationships among DNN weights and is unlikely to be known to the observer/receiver. This suggests that the exact MAP and MMSE estimates are not computable by the observer.

\begin{rem}
Exact characterization of  $p(\bm{w})$ is non-trivial even at the transmitter, especially for NoisyNNs arising from noisy training, where $\bm{w}$ is the weights of a half-trained DNN.
\end{rem}

Nevertheless, we can approximate the prior distribution $p(\bm{w})$ from the sample statistics of $\bm{w}$. Notice that, 1) the parameter vector $\bm{w}$ is a realization of the prior distribution $p(\bm{w})$; 2) once generated (after some training epochs), the parameter vector is determined and can be characterized by its sample statistics.
In this light, we can postulate that the elements of $\bm{w}$ are sampled from a generic Gaussian random variable $W$ in an i.i.d. manner \cite{GaussianAssumption1,GaussianAssumption2}. In particular, the Gaussian is parameterized by the sample mean and sample variance of $\bm{w}$.

Formally, we define a Gaussian random variable $\mathcal{W}\sim\mathcal{N}(\mathcal{W};\mu_w,\sigma^2_w)$, where $\mu_w$ and $\sigma^2_w$ are the sample mean and sample variance of $\bm{w}=\{w[i]:i=1,2,...,d\}$:
\begin{equation*}
 \mu_w =\frac{1}{d}\sum_{i=1}^{d}w[i],~~
 \sigma^2_w = \frac{1}{d}\sum_{i=1}^{d}(w[i]-\mu_w)^2.
\end{equation*}

Given this approximation, the elements of the observed sequence $\bm{r}$ are also i.i.d., and can be viewed as realizations of a random variable $\mathcal{R=W+Z}$, where $\mathcal{Z}\sim\mathcal{N}(\mathcal{Z};0,\sigma^2_z)$ and $\mathcal{R}\sim\mathcal{N}(\mathcal{R};\mu_w,\sigma^2_w+\sigma^2_z)$. Correspondingly, the MMSE and MAP estimates can be written as
\begin{eqnarray*}
\hat{\mathcal{W}}^{\text{MMSE}}
\hspace{-0.2cm}& = &\hspace{-0.2cm} \arg\min_{\hat{\mathcal{W}}} \int \left(\hat{\mathcal{W}}-\mathcal{W}\right)^2 p(\mathcal{R}|\mathcal{W})p(\mathcal{W}) d\mathcal{W}, \\
\hat{\mathcal{W}}^{\text{MAP}}
\hspace{-0.2cm}& = &\hspace{-0.2cm} \arg\max_{\hat{\mathcal{W}}} p(\mathcal{R}|\mathcal{W})p(\mathcal{W}).
\end{eqnarray*}

The multiplication of two Gaussians is still a Gaussian, thus, $p(\mathcal{R},\mathcal{W}) \propto p(\mathcal{R}|\mathcal{W})p(\mathcal{W}) \sim \mathcal{N}(\mathcal{W}; \mu_p, \sigma^2_p)$, where
\begin{equation*}
\mu_p = \frac{\sigma^2_w\mathcal{R} +\mu_w\sigma^2_z}{\sigma^2_w+\sigma^2_z},~~ \sigma^2_p=\frac{\sigma^2_w\sigma^2_z}{\sigma^2_w+\sigma^2_z}.
\end{equation*}
As a result, the MMSE and MAP metrics are equivalent in that maximizing the posterior probability is equivalent to minimizing the MSE when the joint distribution $p(\mathcal{R},\mathcal{W})$ is Gaussian. We shall focus on $\hat{\mathcal{W}}^{\text{MMSE}}$ below.

The MSE between the estimated weight $\hat{\mathcal{W}}$ and the true weight $\mathcal{W}$ can be written as
\begin{equation}\label{eq:MSEw}
\text{MSE}_w = \mathbb{E}\left[\left.\left(\hat{\mathcal{W}}\!-\!\mathcal{W}\right)^2\right|\mathcal{R} \right] = \int\! \left(\hat{\mathcal{W}}\!-\!\mathcal{W}\right)^2 p(\mathcal{W},\mathcal{R}) d\mathcal{W}.
\end{equation}
Differentiating $\text{MSE}_w$ with respect to $\hat{\mathcal{W}}$ gives us
\begin{eqnarray*}
&& \frac{\partial \text{MSE}_w}{\partial \hat{\mathcal{W}}} = \int \left(\hat{\mathcal{W}}-\mathcal{W}\right) p(\mathcal{W},\mathcal{R}) d\mathcal{W} = 0, \\
&&\hspace{0.8cm} \hat{\mathcal{W}} =   \int \mathcal{W}  p(\mathcal{W},\mathcal{R}) d\mathcal{W} = \mu_p.
\end{eqnarray*}

We then arrive at the following MMSE estimator.

\begin{defi}[MMSE estimation]
Given the observed DNN weight vector $\bm{r}\in\RR^d$ in \eqref{eq:r}, an MMSE estimator estimates the uncontaminated weight vector $\bm{w}$ by
\begin{equation}\label{eq:MMSE}
\hat{\bm{w}}^{\text{MMSE}} = \frac{\sigma^2_w}{\sigma^2_w+\sigma^2_z}\bm{r} + \frac{\mu_w\sigma^2_z}{\sigma^2_w+\sigma^2_z}\bm{e},
\end{equation}
where $\bm{e}$ is an all-ones vector.
The MMSE estimate is also the MAP estimate.
\end{defi}

The MMSE estimator in \eqref{eq:MMSE} minimizes the MSE between $\hat{\bm{w}}$ and $\bm{w}$. However, the ultimate goal of denoising is not to minimize the MSE of the estimated weights, but to maximize the inference accuracy of the reconstructed model. For example, consider a regression problem with the MSE loss function. Suppose the DNN weights $\bm{w}$ are well-trained so that $y^n=\mathcal{F}(\bm{x}^n|{\bm{w}})$, the goal of denoising is then to minimize the MSE between the original DNN output and the denoised DNN output. That is,
\begin{eqnarray*}
\hat{\bm{w}}^\text{denoiser}
\hspace{-0.25cm}&=&\hspace{-0.25cm}
\arg\min_{\hat{\bm{w}}}\frac{1}{N}\left(\mathcal{F}(\bm{x}^n|\hat{\bm{w}})-\bm{y}^n \right)^\top\left(\mathcal{F}(\bm{x}^n|\hat{\bm{w}})-\bm{y}^n \right) \\
\hspace{-0.25cm}&\triangleq&\hspace{-0.25cm}
\arg\min_{\hat{\bm{w}}}\text{MSE}_{\mathcal{F}}.
\end{eqnarray*}
Said in another way, the MMSE estimator in \eqref{eq:MMSE}, which minimizes $\text{MSE}_{w}$, does not necessarily minimize $\text{MSE}_{\mathcal{F}}$.

Analytically deriving the optimal denoiser that maximizes the inference accuracy of the reconstructed DNN is non-trivial due to the high nonlinearity of the DNN $\mathcal{F}$. Thus, we resort to empirical approaches and improve the MMSE criterion  by two compensators to reduce the error in the denoised DNN output.

\subsection{Bayesian denoiser with compensators}
There are two empirical facts about DNNs \cite{Pruning1,Pruning2}: i) most of parameter values in a DNN are very small in magnitude; and ii) as far as the inference accuracy is concerned, the parameters with a larger magnitude matter more than that those with a smaller magnitude. If we examine the MSE metric in \eqref{eq:MSEw}, however, each parameter contributes equally to $\text{MSE}_w$ regardless of its magnitude. This implies that the MMSE criterion and our ultimate goal of denoising are mismatched.

Considering that there is a large population of parameters that are very small in magnitude, we propose to add a population compensator to the MSE metric so that the estimation error of larger parameters (larger in magnitude) weighs more than the estimation error of smaller parameters. Specifically, instead of minimizing $\text{MSE}_w$, we aim to minimize $\text{MSE}_p$ as defined below.
\begin{eqnarray}\label{eq:MSEp}
\text{MSE}_p \hspace{-0.2cm}& = &\hspace{-0.2cm} \mathbb{E}\left[\left.\left(\hat{\mathcal{W}}-\mathcal{W}\right)^2e^{\lambda \mathcal{W}^2} \right|\mathcal{R} \right] \nonumber\\
\hspace{-0.2cm}& = &\hspace{-0.2cm} \int \left(\hat{\mathcal{W}}-\mathcal{W}\right)^2 e^{\lambda \mathcal{W}^2} p(\mathcal{W},\mathcal{R}) d\mathcal{W},
\end{eqnarray}
where $\lambda$ is a temperature parameter that controls the extent to which we compensate for the smaller populations of larger parameters.
Based on the new $\text{MSE}_p$ metric, an $\text{MMSE}_p$ denoiser is devised in Proposition \ref{prop:MMSEp}.

\begin{prop}[The $\text{MMSE}_p$ denoiser]\label{prop:MMSEp}
Given the observed noisy weight vector $\bm{r}\in\RR^d$ in \eqref{eq:r}, an $\text{MMSE}_p$ denoiser that minimizes $\text{MSE}_p$ reconstructs the weight vector $\bm{w}$ by
\begin{equation}\label{eq:MMSEp}
\hat{\bm{w}}^{\text{MMSE}_p} =
\frac{\sigma^2_w}{\sigma^2_w+(1-2\sigma^2_w\lambda)\sigma^2_z}\bm{r}+\frac{\mu_w\sigma^2_z}{\sigma^2_w+(1-2\sigma^2_w\lambda)\sigma^2_z}\bm{e},
\end{equation}
where the temperature parameter $0\leq\lambda<\frac{1}{2\sigma^2_w}+\frac{1}{2\sigma^2_z}$.
\end{prop}

\begin{NewProof}
Let $q_\lambda(\mathcal{W},\mathcal{R})=e^{\lambda \mathcal{W}^2}p(\mathcal{W},\mathcal{R})$. Since $p(\mathcal{W},\mathcal{R})\sim\mathcal{N}(\mathcal{W}; \mu_p, \sigma^2_p)$, we have
\begin{eqnarray*}
q_\lambda(\mathcal{W},\mathcal{R})\propto e^{\lambda \mathcal{W}^2} e^{-\frac{(\mathcal{W}-\mu_p)^2}{2\sigma^2_p}} \propto   e^{-\frac{(\mathcal{W}-\mu_{\lambda})^2}{2\sigma^2_{\lambda}}},
\end{eqnarray*}
where
\begin{eqnarray*}
\mu_{\lambda}
\hspace{-0.2cm}& = &\hspace{-0.2cm} \frac{\mu_p}{1-2\sigma^2_p\lambda} = \frac{\sigma^2_w\mathcal{R}+\mu_w\sigma^2_z}{\sigma^2_w+(1-2\sigma^2_w\lambda)\sigma^2_z}, \\
\sigma^2_{\lambda}
\hspace{-0.2cm}& = &\hspace{-0.2cm} \frac{\sigma^2_p}{1-2\sigma^2_p\lambda} = \frac{\sigma^2_w\sigma^2_z}{\sigma^2_w+(1-2\sigma^2_w\lambda)\sigma^2_z}.
\end{eqnarray*}
In other words, $q_\lambda(\mathcal{W},\mathcal{R})$ is also Gaussian:
\begin{eqnarray}\label{eq:qlamb}
q_\lambda(\mathcal{W},\mathcal{R})\sim\mathcal{N}(\mathcal{W};\mu_{\lambda},\sigma^2_{\lambda}).
\end{eqnarray}
In particular, to ensure that $\sigma^2_{\lambda}>0$, we impose $\lambda<\frac{1}{2\sigma^2_w}+\frac{1}{2\sigma^2_z}$.

Substituting \eqref{eq:qlamb} into \eqref{eq:MSEp} suggests that minimizing $\text{MSE}_p$ is equivalent to minimizing $\text{MSE}_w$ with a modified prior Gaussian with mean $\mu_{\lambda}$ and variance $\sigma^2_{\lambda}$ (as opposed to $\mu_p$ and $\sigma^2_p$). Following \eqref{eq:MSEw}, the estimate $\hat{\mathcal{W}}$ that minimizes $\text{MSE}_p$ is then
\begin{eqnarray*}
\hat{\mathcal{W}} =  \int \mathcal{W}  q_{\lambda}(\mathcal{W},\mathcal{R}) d\mathcal{W} = \mu_{\lambda},
\end{eqnarray*}
which gives us \eqref{eq:MMSEp}.
\end{NewProof}

\begin{figure}[t]
  \centering
  \includegraphics[width=0.75\linewidth]{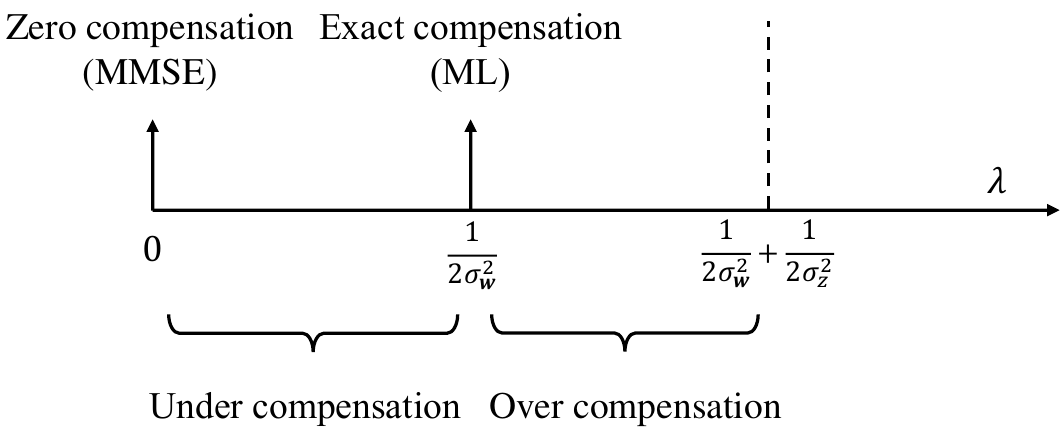}\\
  \caption{Relations among the ML, MMSE, and $\text{MMSE}_p$ estimators.}
\label{fig:2}
\end{figure}

Fig.~\ref{fig:2} depicts a comparison among the ML estimator, the MMSE estimator, and the $\text{MMSE}_p$ denoiser, where we set $\mu_w=0$ because the sample mean of the DNN weights is typically very small. As can be seen, for the $\text{MMSE}_p$ denoiser, the extent to which the population is compensated is controlled by the temperature parameter $\lambda$. For different values of $\lambda$, the ML and MMSE estimators can be viewed as special cases of the $\text{MMSE}_p$ denoiser. Specifically,
\begin{enumerate}[leftmargin=0.5cm]
\item When $\lambda=0$, we have $\hat{\bm{w}}^{\text{MMSE}_p}=\hat{\bm{w}}^{\text{MMSE}}$, and the $\text{MMSE}_p$ denoiser reduces to the MMSE estimator. We call it the zero-compensation point. For this setting, the contribution to the MSE of a parametric value $w[i]$ is proportional to $p(\mathcal{W}=w[i])$. Thus, the more likely $w[i]$ is, the more the corresponding estimation error counts toward the MSE. Given that small weights are more likely, this setting may over-value the importance of small weights toward our denoiser.
\item When $\lambda=\frac{1}{2\sigma^2_w}$, we have $\hat{\bm{w}}^{\text{MMSE}_p}=\hat{\bm{w}}^{\text{ML}}$, the $\text{MMSE}_p$ denoiser reduces to the ML estimator. We call it the exact-compensation point. In this setting, $p(\mathcal{W})$ does not count. All values of $\bm{w}$ count equally toward the MSE regardless of the relative populations of different $w[i]$ as indicated in $p(\mathcal{W})$. Thus, the population bias in $p(\mathcal{W})$ is compensated away exactly.
\item When $0<\lambda<\frac{1}{2\sigma^2_w}$, we call it the under-compensation region.
\item When $\frac{1}{2\sigma^2_w}<\lambda<\frac{1}{2\sigma^2_w}+\frac{1}{2\sigma^2_z}$, we call it the over-compensation region.
\end{enumerate}

In addition to the population compensator, we have empirically observed that an extra compensation term, dubbed the bias compensator, can further improve the denoising performance. With the bias compensator, our goal is to minimize
\begin{eqnarray}\label{eq:MSEpb}
\text{MSE}_{pb}
\hspace{-0.2cm}& = &\hspace{-0.2cm} \mathbb{E}\left[\left.\left(\hat{\mathcal{W}}-\mathcal{W}\right)^2e^{\lambda \mathcal{W}^2 + \beta \mathcal{W}} \right|\mathcal{R} \right] \nonumber\\
\hspace{-0.2cm}& = &\hspace{-0.2cm}
\int \left(\hat{\mathcal{W}}-\mathcal{W}\right)^2 e^{\lambda \mathcal{W}^2+ \beta \mathcal{W}} p(\mathcal{W},\mathcal{R}) d\mathcal{W},
\end{eqnarray}
where $\beta$ is another temperature parameter to be tuned. This shapes our final design of the $\text{MMSE}_{pb}$ denoiser.

\begin{prop}[The $\text{MMSE}_{pb}$ denoiser]\label{thm:MMSEpb}
Given the observed DNN weight vector $\bm{r}\in\RR^d$ in \eqref{eq:r}, an $\text{MMSE}_{pb}$ denoiser that minimizes $\text{MSE}_{pb}$ reconstructs the weight vector $\bm{w}$ by
\begin{equation}\label{eq:MMSEpb}
\hat{\bm{w}}^{\text{MMSE}_{pb}} = \frac{\sigma^2_w}{\sigma^2_w+(1-2\sigma^2_w\lambda)\sigma^2_z}\bm{r}+\frac{\sigma^2_w\sigma^2_z\beta}{\sigma^2_w+(1-2\sigma^2_w\lambda)\sigma^2_z}\bm{e},
\end{equation}
where $\lambda$ and $\beta$ are temperature parameters, and $0\leq\lambda<\frac{1}{2\sigma^2_w}+\frac{1}{2\sigma^2_z}$.
\end{prop}

\begin{NewProof}
Let $q_{\lambda,\beta}(\mathcal{W},\mathcal{R})=e^{\lambda \mathcal{W}^2+ \beta \mathcal{W}} p(\mathcal{W},\mathcal{R})$. Since $p(\mathcal{W},\mathcal{R})\allowbreak\sim\allowbreak\mathcal{N}(\mathcal{W};\allowbreak \mu_p, \sigma^2_p)$, we have
\begin{eqnarray}\label{eq:qlambbeta}
q_{\lambda,\beta}(\mathcal{W},\mathcal{R})\sim\mathcal{N}(\mathcal{W};\mu_{\lambda,\beta},\sigma^2_{\lambda,\beta}),
\end{eqnarray}
where
\begin{eqnarray*}
\mu_{\lambda,\beta}
\hspace{-0.2cm}& = &\hspace{-0.2cm}
\frac{\mu_p+\sigma^2_p\beta}{1-2\sigma^2_p\lambda} = \frac{\sigma^2_w\mathcal{R}+\sigma^2_z\mu_w+\sigma^2_w\sigma^2_z\beta}{\sigma^2_w+(1-2\sigma^2_w\lambda)\sigma^2_z}, \\
\sigma^2_{\lambda,\beta}
\hspace{-0.2cm}& = &\hspace{-0.2cm}
\frac{\sigma^2_p}{1-2\sigma^2_p\lambda} = \frac{\sigma^2_w\sigma^2_z}{\sigma^2_w+(1-2\sigma^2_w\lambda)\sigma^2_z}.
\end{eqnarray*}

As in \eqref{eq:qlamb}, minimizing $\text{MSE}_{pb}$ is equivalent to minimizing $\text{MSE}_w$ with a modified prior Gaussian with mean $\mu_{\lambda,\beta}$ and variance $\sigma^2_{\lambda,\beta}$ given in \eqref{eq:qlambbeta}. From \eqref{eq:MSEw}, the estimate $\hat{\mathcal{W}}$ that minimizes $\text{MSE}_{pb}$ is then $\mu_{\lambda,\beta}$. Setting $\mu_w=0$ gives us the $\text{MMSE}_{pb}$ denoiser in \eqref{eq:MMSEpb}.

Note that the bias compensator does not change the variance, i.e., $\sigma^2_{\lambda,\beta}=\sigma^2_{\lambda}$. To ensure that $\sigma^2_{\lambda,\beta}>0$, the constraint is still $\lambda<\frac{1}{2\sigma^2_w}+\frac{1}{2\sigma^2_z}$.
\end{NewProof}

In our $\text{MMSE}_{pb}$ denoiser, there are four parameters to be determined: the sample variance of DNN weights $\sigma^2_w$, the noise variance $\sigma^2_z$, and the temperature parameters $\lambda$ and $\beta$. The first two parameters $\sigma^2_w$ and $\sigma^2_z$ are often readily available to an observer/receiver. For example, in wireless communication applications, the noise variance $\sigma^2_z$ can be estimated by the receiver when there is no signal transmission. The sample variance of DNN weights $\sigma^2_w$, on the other hand, can be estimated by the receiver from the sample variance of the observed vector, i.e., $\sigma^2_r$, when there are signal transmissions. This is because $\mu_w=\mu_r$ and $\sigma^2_w=\sigma^2_r-\sigma^2_z$, where
\begin{eqnarray*}
\mu_r =\frac{1}{d}\sum_{i=1}^{d}r[i],~~
\sigma^2_r = \frac{1}{d}\sum_{i=1}^{d}(r[i]-\mu_r)^2,
\end{eqnarray*}
can be obtained directly from the re\-cei\-ved/ob\-ser\-ved samples.

On the other hand, determining the optimal temperature parameters $\lambda$ and $\beta$ can be more intricate.
For small-scale problems, such as the two instances considered in the next section, the optimal temperature parameters can be analytically derived or approximated.
In contrast, for advanced DNNs, such as the convolutional neural network (CNN) and the att\-en\-tion-me\-cha\-nism-ba\-sed neural network considered in Section~\ref{sec:5}, analytically deriving the optimal $\lambda$ and $\beta$ is formidable. 
In that case, we will identify good temperature parameters by a grid search on the validation dataset.

\begin{rem}[Layerwise denoising]\label{rem:layerwise}
In Proposition~\ref{thm:MMSEpb}, the observed noisy vector $\bm{r}$ is formed by all the weights in a DNN; hence, the DNN is denoised as a whole.
Alternatively, we can denoise the DNN layer-by-layer with the $\text{MMSE}_{pb}$ denoiser, in which case the vector $\bm{r}$ consists of the weights in a single larger of the DNN, and $\sigma^2_w$, $\lambda$, and $\beta$ are layer-dependent. 
With layer-wise denoising, the performance can be improved by much, as will be shown in Section \ref{sec:5}, thanks to the additional freedom brought by layer-dependent temperature parameters. The downside, however, is that the grid search becomes more computationally demanding as there are more temperature parameters to be determined.
\end{rem}

\section{Superiority of the $\text{MMSE}_{pb}$ Denoiser}\label{sec:4}
Compared with the widely-used ML estimator, minimizing $\text{MSE}_{pb}$ leads to a simple linear denoiser in \eqref{eq:MMSEpb}. In this section, we will demonstrate the superiority of the $\text{MMSE}_{pb}$ denoiser over the  ML estimation in two small-scale problems, wherein the neural networks $\mathcal{F}$ are assumed to be a quadratic function and a shallow feedforward neural network, respectively.
To simplify notations, throughout this section we write the $\text{MMSE}_{pb}$ denoiser in \eqref{eq:MMSEpb} as $\hat{\bm{w}}^{\text{MMSE}_{pb}}\triangleq \theta(\lambda)\bm{r}+\rho(\lambda,\beta)$, where $\theta(\lambda)$ and $\rho(\lambda,\beta)$ are the multiplicative and additive factors of \eqref{eq:MMSEpb}, respectively.

\subsection{A quadratic neural network function}
In this subsection, we consider the neural network $\mathcal{F}$ to be a quadratic function, as \cite{Schaul} and \cite{Lookahead}.

\begin{defi}\label{defi:quadratic}
The quadratic neural network function $\mathcal{F}$ is defined as
\begin{equation}\label{eq:quadratic}
    \mathcal{F}: y(\bm{x,w})=(\bm{x}-c\bm{e})^\top \bm{W}^2 (\bm{x}-c\bm{e}),
\end{equation}
where
$\bm{x}\in\RR^d$;
$c$ is a constant;
$\bm{e}$ is an all-ones vector;
$\bm{W}$ is a diagonal weight matrix, the diagonal elements of which form the weight vector $\bm{w}\in\RR^d$.
\end{defi}

When the weight vector is contaminated by AWGN, the weight matrix $\bm{W}$ in \eqref{eq:quadratic} is replaced by $\bm{R=W+Z}$, where $\bm{Z}$ is a diagonal matrix with diagonal elements being the noise vector $\bm{z}$: $z_i \sim \mathcal{N}(0,\sigma_{z}^2)$.

After $\text{MMSE}_{pb}$ denoising, the neural network output can be written as
\begin{equation} \tilde{y}\big(\bm{x,w,z},\theta(\lambda),\rho(\lambda,\beta)\big)=(\bm{x}-c\bm{e})^\top \bm{W}^2_{pb} (\bm{x}-c\bm{e}),
\end{equation}
where $\bm{W}_{pb}=\theta(\lambda)\bm{R}+\rho(\lambda,\beta)\bm{I}_{d\times d}$ is a diagonal matrix with diagonal elements being the denoised vector $\hat{\bm{w}}^{\text{MMSE}_{pb}}$.
In particular, when $\lambda=\frac{1}{2\sigma^2_w}$ and $\beta=0$,  $\tilde{y}\big(\bm{x},\allowbreak\bm{w},\allowbreak\bm{z},\allowbreak\theta(\frac{1}{2\sigma^2_w})=1,\allowbreak\rho(\frac{1}{2\sigma^2_w},0)=0\big)$ is the neural network output with the ML estimation, as shown in Fig.~\ref{fig:2}.

To evaluate the performance of the denoiser, next we define an error measurement for the denoised neural network.

\begin{defi}[Squared output error]\label{defi:error}
The performance of a denoiser is measured by the squared error between the noiseless neural network and the denoised neural network outputs:
\begin{equation}
\mathcal{D}=\Big[\tilde{y}\big(\bm{x,w,z},\theta(\lambda),\rho(\lambda,\beta)\big)-y(\bm{x,w})\Big]^2.
\end{equation}
For a given NoisyNN, the optimal temperature parameters $\lambda^*$ and $\beta^*$ minimize the expected squared error $\bar{\mathcal{D}}\triangleq\mathbb{E}_{\bm{x,w,z}} \mathcal{D}$, i.e.,
\begin{equation*}
\lambda^*,\beta^*=\arg\min_{\lambda,\beta}\bar{\mathcal{D}}.
\end{equation*}
\end{defi}

\begin{thm}[Optimal $\text{MMSE}_{pb}$ denoiser]\label{thm:exp1}
Consider the qua\-dratic neural network function in Definition \ref{defi:quadratic}.
Suppose the elements of the input vector $\bm{x}$ are i.i.d. and follow the uniform distribution in $[-1,1)$, i.e., $x_i \sim \allowbreak\mathcal{U}(-1,\allowbreak 1)$; the elements of $\bm{w}$ follow a zero-mean i.i.d. Gaussian distribution, i.e., $w_i \sim\allowbreak \mathcal{N}(0,\allowbreak\sigma_{\bm{w}}^2)$. Then, the optimal $\lambda$ and $\beta$ that minimize the expected squared error $\bar{\mathcal{D}}$ are given by
\begin{equation}\label{eq:optimalpara}
\lambda^*=\frac{1}{2\sigma^2_w}+\frac{1}{2\sigma^2_z}-\sqrt{\frac{C^\prime_1\sigma^2_w}{C^\prime_1\sigma^2_w+C^\prime_2\sigma^2_z}}\frac{\sigma^2_w+\sigma^2_z}{2\sigma^2_w\sigma^2_z}~\text{and}~\beta^*=0,
\end{equation}
where the constants $C^\prime_1\triangleq(c^2-\frac{1}{3})^2d^2+2(c^4+\frac{10}{3}c^2+\frac{11}{45})d$ and $C^\prime_2\triangleq(c^2-\frac{1}{3})^2d^2+\frac{4}{3}(2c^2+\frac{1}{15})d$.

Compared with the ML estimator, the optimal $\text{MMSE}_{pb}$ denoiser reduces the expected squared error $\bar{\mathcal{D}}$ by a factor of
\begin{equation}\label{eq:factor}
\frac{\bar{\mathcal{D}}^{\text{ML}}\!-\!\bar{\mathcal{D}}^{\text{MMSE}_{pb}}}{\bar{\mathcal{D}}^{\text{ML}}}
\!=\!\frac{(2C^\prime_1\sigma^2_w-C^\prime_2\sigma^2_w+C^\prime_1\sigma^2_z)^2\sigma^2_z}{C^\prime_1(\sigma^2_w+\sigma^2_z)^2(2C^\prime_1\sigma^2_w-2C^\prime_2\sigma^2_w+C^\prime_1\sigma^2_z)}.
\end{equation}
\end{thm}

\begin{NewProof}
(Sketch, see Appendix \ref{sec:AppA} for the detailed proof).
Given the quadratic neural network function in Definition \ref{defi:quadratic}, we first derive the closed-form expected error $\bar{\mathcal{D}}$  as a function of $\theta$ and $\rho$ for the $\text{MMSE}_{pb}$ denoiser:
\begin{eqnarray}\label{eq:expbarD}
\bar{\mathcal{D}} &&\hspace{-0.55cm} = C_1^{\prime}\big[\theta^4(\sigma_{\bm{w}}^2+\sigma_{\bm{z}}^2)^2-2\theta^2\sigma_{\bm{w}}^4 + 2\theta^2\rho^2(\sigma_{\bm{w}}^2+\sigma_{\bm{z}}^2)+\sigma_{\bm{w}}^4\big] \nonumber\\
&&\hspace{-0.5cm} 
+ C_2^{\prime}\left(\rho^4-2\theta^2\sigma_{\bm{w}}^2\sigma_{\bm{z}}^2-2\sigma_{\bm{w}}^2\rho^2\right),
\end{eqnarray}
where $C_1^{\prime}$ and $C_2^{\prime}$ are as defined in \eqref{eq:optimalpara}.

$\bar{\mathcal{D}}$ is an even function of both $\theta$ and $\rho$. Thus, we focus on the region  $\begin{Bmatrix}\theta\geq0,\rho\geq0\end{Bmatrix}$ and examine the monotonicity of $\bar{\mathcal{D}}$. It turns out that $\bar{\mathcal{D}}$ has four critical points in this region.
\begin{eqnarray}
\label{eq:P1}
&&\hspace{-0.5cm} \text{P1}: \theta=0,~\rho=0; \\
&&\hspace{-0.5cm} \text{P2}: \theta=0,~\rho=\sigma_{\bm{w}}; \nonumber\\
\label{eq:P3}
&&\hspace{-0.5cm} \text{P3}: \theta=\frac{\sigma^2_w}{\sigma^2_w+\sigma^2_z}\sqrt{1+\frac{C^\prime_2\sigma^2_z}{C^\prime_1\sigma^2_w}},~\rho=0;~\text{and} \\
&&\hspace{-0.5cm} \text{P4}: \theta=\sqrt{\frac{C_2^{\prime}}{C_1^{\prime}}}\frac{\sigma_{\bm{w}}\sigma_{\bm{z}}}{\sigma_{\bm{w}}^2+\sigma_{\bm{z}}^2},~\rho=\frac{\sigma_{\bm{w}}^2}{\sqrt{\sigma_{\bm{w}}^2+\sigma_{\bm{z}}^2}}. \nonumber
\end{eqnarray}

Examining the positive definiteness of the Hessian of $\bar{\mathcal{D}}$ at the four critical points indicates that P1 is a local maximum, while P2, P3, and P4 are local minima. It can be shown that P3 is the global minimum. By symmetry, another global minimum is
\begin{equation}\label{eq:P3prime}
\text{P3}^\prime: \theta=-\frac{\sigma^2_w}{\sigma^2_w+\sigma^2_z}\sqrt{1+\frac{C^\prime_2\sigma^2_z}{C^\prime_1\sigma^2_w}},~\rho=0. 
\end{equation}

From \eqref{eq:MMSEpb}, an additional constraint of $\theta$ is $\frac{\sigma^2_w}{\sigma^2_w+\sigma^2_z}\leq\theta<\infty$ since $0\leq\lambda<\frac{1}{2\sigma^2_w}+\frac{1}{2\sigma^2_z}$. Thus, P3 is the only global minimum in the region of interest.
The optimal $\lambda^*$ and $\beta^*$ in \eqref{eq:optimalpara} and the performance gain of the $\text{MMSE}_{pb}$ denoiser in \eqref{eq:factor} can then be derived.
\end{NewProof}

\begin{figure}[t]
\centering
\subfigure{
\includegraphics[width=0.7\linewidth]{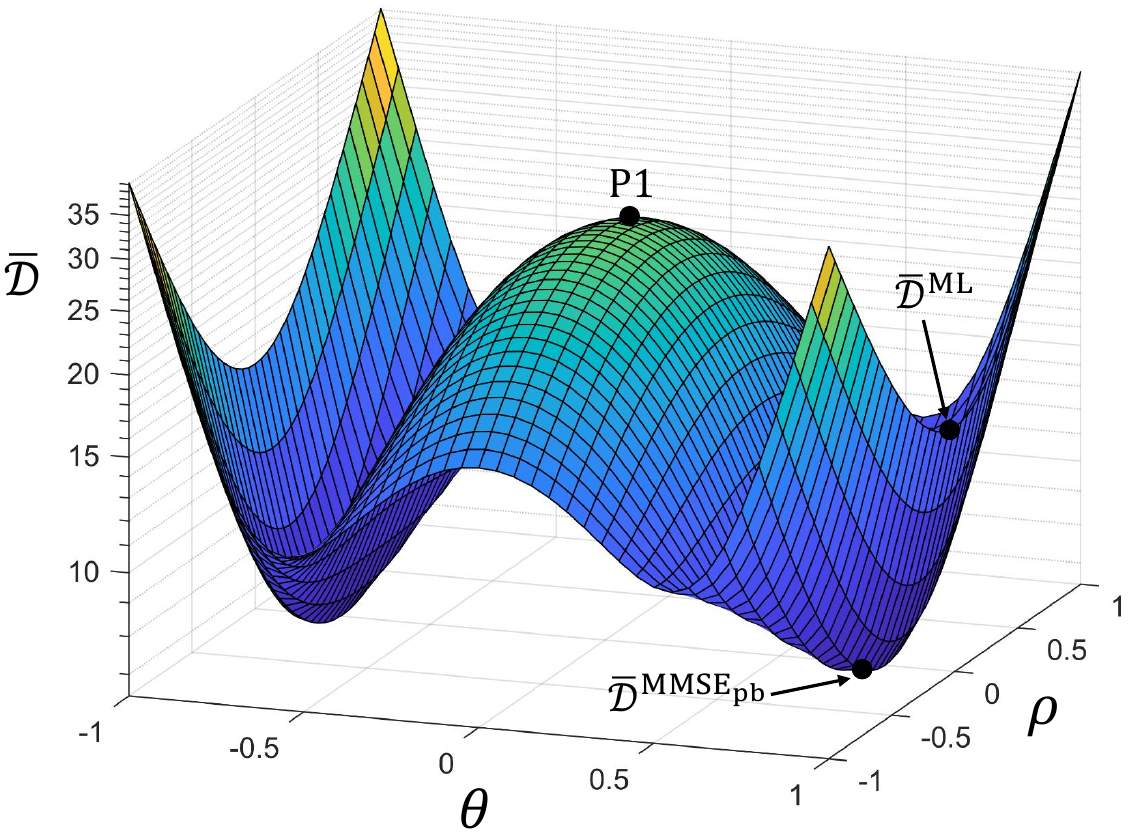}
}
\quad
\subfigure{
\includegraphics[width=0.7\linewidth]{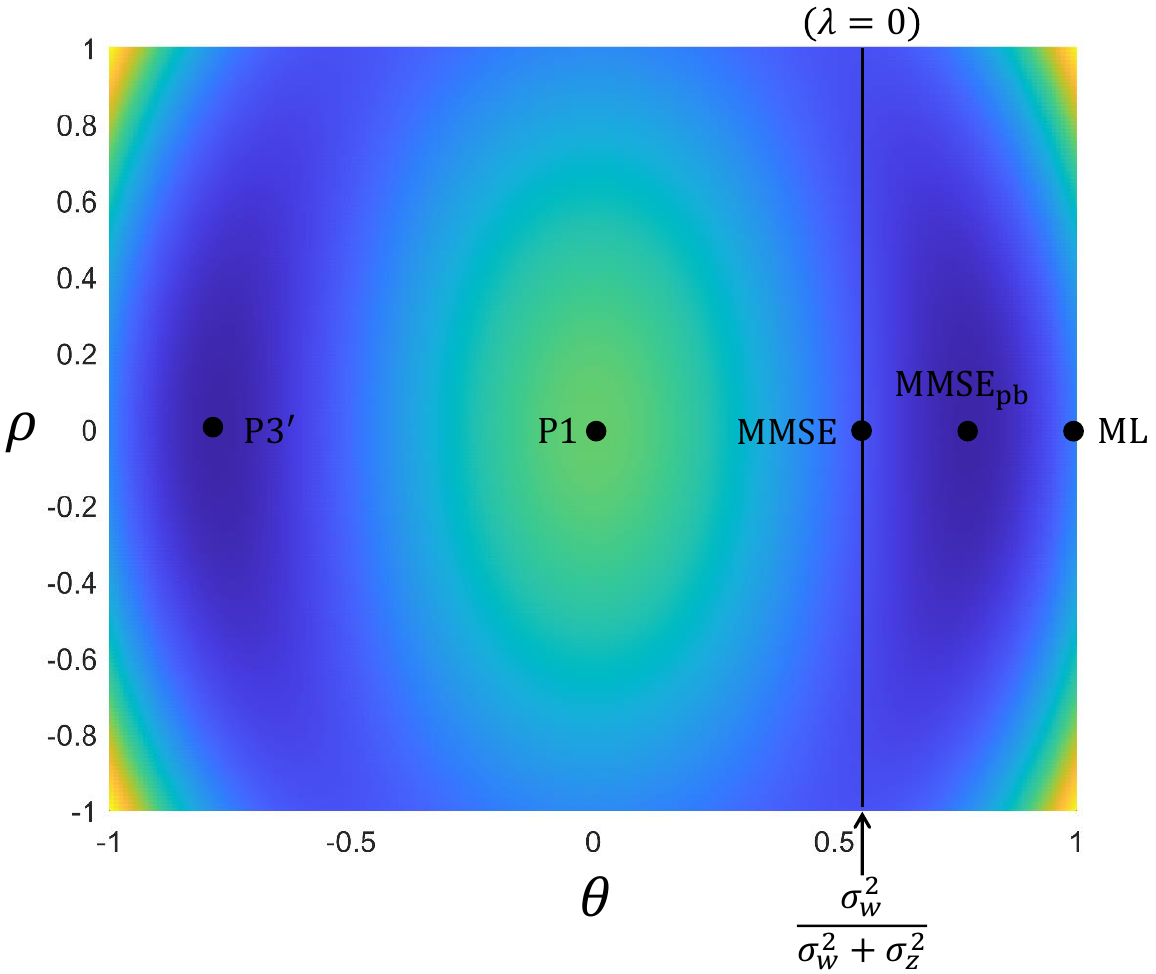}
}
\caption{An illustration of Theorem \ref{thm:exp1}, where $\sigma^2_w=2.25$, $\sigma^2_z=1$, $d=5$, and $c=0.1$. The lower figure is a contour plot of the upper figure.}
\label{fig:exp1}
\end{figure}

To illustrate Theorem \ref{thm:exp1}, an example is presented in Fig.~\ref{fig:exp1}, wherein we set $\sigma^2_w=2.25$, $\sigma^2_z=1$, $d=5$, and $c=0.1$. The upper figure presents the shape of the expected squared error $\bar{\mathcal{D}}$ in \eqref{eq:expbarD} as a function of $\theta$ and $\rho$. The lower figure is a contour plot of the upper figure. As predicted, P1 in \eqref{eq:P1} is a local maximum; $\text{P3}$ and $\text{P3}^\prime$ are global minima. In the region of interest $\theta\geq\frac{\sigma^2_w}{\sigma^2_w+\sigma^2_z}$ (i.e., $\lambda\geq 0$), the $\text{MMSE}_{pb}$ denoiser minimizes the expected squared error of the network output. Compared with the ML estimation, the expected squared error is reduced by $58\%$, as predicted in \eqref{eq:factor}.

\subsection{A shallow feedforward neural network}
In this subsection, we consider a three-layer feedforward neural network with one input neuron, $N$ hidden neurons, and one output neuron, as in \cite{An1996}.

\begin{figure}[t]
  \centering
  \includegraphics[width=0.85\linewidth]{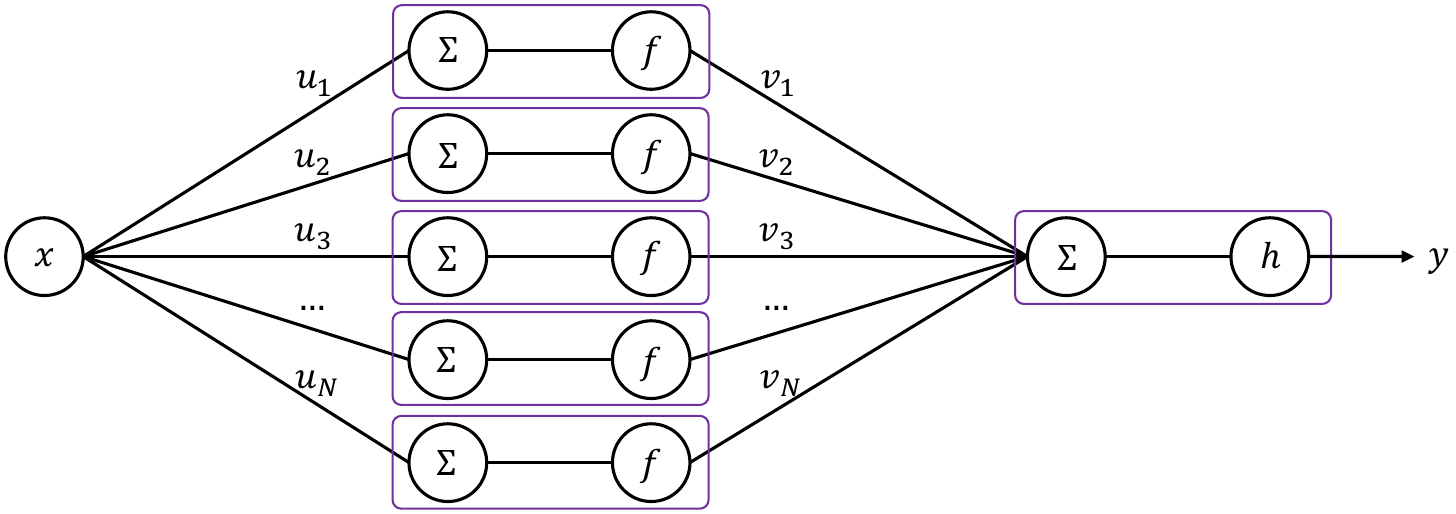}\\
  \caption{A three-layer feedforward neural network with one input neural, $N$ hidden neurons, and one output neuron.}
\label{fig:exp2}
\end{figure}

\begin{defi}\label{defi:exp2}
The three-layer feedforward neural network is defined as follows:
\begin{itemize}
\item The input of the neural network is $x \in \mathbb{R}$;
\item The neural network weights between the input and hidden neurons are $\bm{u}=[u_1,u_2,...,u_N]^\top$;
\item The hidden neurons have no bias. The activation function $f$ is $\tanh$. Note that $d\tanh(x)/dx = 1 - \tanh^2(x)$;
\item The neural network weights between the hidden neurons and the output neuron are $\bm{v}=[v_1,v_2,...,v_N]^\top$;
\item The output neuron has no bias. The activation function is $h(x)=x$;
\item The output of the neural network is $y \in \mathbb{R}$.
\end{itemize}
The neural network function can be written as
\begin{eqnarray}
\mathcal{F}:y(x,\bm{w})=\sum_{i=1}^{N}v_i\tanh(u_ix),
\end{eqnarray}
where the weight vector $\bm{w}\in\mathbb{R}^d$ is formed by $\bm{u}$ and $\bm{v}$, i.e., $\bm{w}=[\bm{u}^\top,\bm{v}^\top]^\top$ and $d=2N$.
\end{defi}

The architecture of the three-layer neural network is shown in Fig.~\ref{fig:exp2}.
When passing through a noisy medium, the network weights suffer from random perturbations. The NoisyNN can be written as
\begin{eqnarray}
y(x,\bm{w}+\bm{z})=\sum_{i=1}^{N}(v_i+\Delta v_i)\tanh\big((u_i+\Delta u_i)x\big),
\end{eqnarray}
where $\bm{z}=[\bm{\Delta u}^\top,\allowbreak\bm{\Delta v}^\top]^\top$ and $\bm{\Delta u}=[\Delta u_1,\allowbreak\Delta u_2,\allowbreak ...,\allowbreak\Delta u_N]^\top$, $\bm{\Delta v}=[\Delta v_1,\allowbreak\Delta v_2,...,\allowbreak\Delta v_N]^\top$ are AWGN noise vectors -- the elements of $\bm{\Delta u}$ and $\bm{\Delta v}$ are sampled from a Gaussian distribution $\mathcal{N}(0,\sigma_{\bm{z}}^2)$ in an i.i.d. manner.

After $\text{MMSE}_{pb}$ denoising, the neural network output can be written as
\begin{eqnarray}\label{eq:denoised2}
&&\hspace{-0.5cm}\tilde{y}(x,\bm{w},\bm{z},\theta(\lambda),\rho(\lambda,\beta))= \\
&&\hspace{0.7cm}\sum_{i=1}^{N}[\theta v_i+\theta \Delta v_i+\rho]\tanh(\theta u_ix+\theta \Delta u_ix+\rho x), \nonumber
\end{eqnarray}
where $\theta(\lambda)$ and $\rho(\lambda,\beta)$ are the multiplicative and additive factors in (13), respectively. Setting $\lambda=1/2\sigma_{\bm{w}}^2$ and $\beta=0$ gives us the ML estimation.

Neural networks are often difficult to analyze due to the non-linear activation functions (e.g., $\tanh$, $\text{sigmoid}$ and $\text{ReLU}$). To analytically derive the expected output error $\bar{\mathcal{D}}=\mathbb{E}_{x,\bm{w},\bm{z}}\mathcal{D}$ for the denoised neural network in \eqref{eq:denoised2}, we first approximate the denoised network output by Taylor series expansion in Lemma~\ref{lem:app}.

To ease the exposition, we introduce the following notations:
\begin{eqnarray*}
g_i                 \hspace{-0.2cm}& \triangleq &\hspace{-0.2cm} \tanh(\theta u_ix+\rho x),\\
g_i^{\prime}        \hspace{-0.2cm}& \triangleq &\hspace{-0.2cm} 1-{\tanh}^2(\theta u_ix+\rho x),\\
g_i^{\prime \prime} \hspace{-0.2cm}& \triangleq &\hspace{-0.2cm} {\tanh}^3(au_ix+bx)-\tanh(\theta u_ix+\rho x).
\end{eqnarray*}
Note that
\begin{eqnarray*}
\frac{\partial g_i}{\partial u_i}=\theta x g_i^{\prime},~~\frac{\partial g_i^{\prime}}{\partial u_i}=2\theta xg_i^{\prime \prime}.
\end{eqnarray*}

\begin{lem}[Approximating the denoised NoisyNN]\label{lem:app}
For a given neural network with weight vector $\bm{w}$ and input $x$, the expected output of the denoised neural network in \eqref{eq:denoised2} can be approximated by
\begin{eqnarray}\label{eq:lemma1}
\mathbb{E}_{\bm{z}}\tilde{y} \approx \tilde{y}_0+\sigma_{\bm{z}}^2\left(\sum_{i=1}^{N}(\theta v_i+\rho)\theta^2 x^2 g_i^{\prime \prime}\right),
\end{eqnarray}
where $\tilde{y}_0 \triangleq \tilde{y}(x,\bm{w},\bm{0},\theta,\rho)$.
\end{lem}

\begin{NewProof}
Notice that in \eqref{eq:denoised2}, $\bm{z}$ is a perturbation on $\bm{w}$. For a given $\bm{w}$ and $x$, the expected output of \eqref{eq:denoised2}, i.e., $\mathbb{E}_{\bm{z}}\tilde{y}$, can be approximated by the Taylor series expansion of $\tilde{y}$ at $\bm{z}=\bm{0}$. This gives us
\begin{eqnarray*}
\tilde{y}=\tilde{y}_0+\bm{z}^T \frac{\partial \tilde{y}_0}{\partial \bm{w}}+\frac{1}{2} \bm{z}^T \frac{\partial^2 \tilde{y}_0}{\partial \bm{w}^2} \bm{z}+\cdots,
\end{eqnarray*}
where $\tilde{y}_0$ is short for $\tilde{y}(x,\bm{w},\bm{0},\theta,\rho)$.

Taking expectation on both sides of \eqref{eq:lemma1} over $\bm{z}$, all the even terms of the right-hand side (RHS) vanish, because the odd-order moments of a zero-mean Gaussian (e.g., AWGN) are 0. Therefore, we have
\begin{eqnarray}\label{eq:lemma2}
\mathbb{E}_{\bm{z}}\tilde{y}
\hspace{-0.2cm}&\approx&\hspace{-0.2cm}
\tilde{y_0} +\frac{1}{2}\bm{z}^T \frac{\partial^2 \tilde{y_0}}{\partial \bm{w}^2}\bm{z} \nonumber\\
\hspace{-0.2cm}&\overset{(a)}{=}&\hspace{-0.2cm}
\tilde{y_0}+\frac{\sigma_{\bm{z}}^2}{2}\left(\sum_{i=1}^{N} \frac{\partial^2 \tilde{y_0}}{\partial v_i^2}+\sum_{i=1}^{N} \frac{\partial^2 \tilde{y_0}}{\partial u_i^2}\right),
\end{eqnarray}
where we have ignored the higher-order terms of the Taylor series expansion in the approximation; step (a) follows because $\mathbb{E}(z_i z_j)=\sigma_{\bm{z}}^2 \delta(i-j)$. That is, the off-diagonal elements of the matrix ${\partial^2 \tilde{y_0}}/{\partial \bm{w}^2}$ vanish because of the Independence of the noise terms, and hence, only the diagonal elements retain.

Further, we have
\begin{eqnarray}
\label{eq:lemma3}
\frac{\partial \tilde{y}_0}{\partial v_i} = \theta g_i,&&\hspace{-0.5cm}
\frac{\partial^2 \tilde{y}_0}{\partial v_i^2} = 0; \\
\label{eq:lemma4}
\frac{\partial \tilde{y}_0}{\partial u_i} = (\theta v_i+\rho)\theta x g_i^{\prime}, &&\hspace{-0.5cm}
\frac{\partial^2 \tilde{y}_0}{\partial u_i^2} = 2(\theta v_i+\rho)\theta^2 x^2 g_i^{\prime \prime}.
\end{eqnarray}

Substituting \eqref{eq:lemma3} and \eqref{eq:lemma4} into \eqref{eq:lemma2} gives us \eqref{eq:lemma1}.
\end{NewProof}

\begin{thm}[Optimal $\text{MMSE}_{pb}$ denoiser]\label{thm:exp2}
For the three-layer neural network in Definition \ref{defi:exp2}, suppose that the input $x$ follows uniform distribution over $[-c,c)$, i.e., $x\sim \mathcal{U}(-c,c)$, and the elements of $\bm{w}=[\bm{u}^\top,\bm{v}^\top]^\top$ follow an i.i.d. Gaussian distribution, i.e., $u_i,v_i \sim \mathcal{N}(0,\sigma_{\bm{w}}^2)$.
The expected output error $\bar{{\mathcal{D}}}$ of the denoised neural network can be approximated by
\begin{eqnarray}\label{eq:exp2_D}
&&\hspace{-0.5cm} \bar{{\mathcal{D}}} \approx
\frac{N}{3} c^2 \sigma_{\bm{w}}^4 \theta^4 - 
\frac{2N}{3}c^2 \sigma_{\bm{w}}^4 \theta^2 + 
\frac{2N}{3} c^2 (\sigma_{\bm{w}}^2 +\sigma_{\bm{z}}^2) \theta^2 \rho^2 + \nonumber\\
&&\hspace{-0.35cm}
\frac{2N(N-1)}{3} \sigma_{\bm{w}}^2 \theta^2 \rho^6 +
\frac{2N}{3} c^2 \sigma_{\bm{w}}^2 \sigma_{\bm{z}}^2 \theta^4
+ N(N-1) \sigma_{\bm{w}}^4 \theta^4 \rho^4 \nonumber\\
&&\hspace{-0.35cm}
+\frac{N^2}{3} c^2 \rho^4 + \frac{N(N-1)}{9} \rho^8 + \frac{N}{3} c^2 \sigma_{\bm{w}}^4 + \psi(c^4).
\end{eqnarray}
In particular, $\psi(c^4)$ denotes a polynomial in which the order of $c$ in each term of $\psi(c^4)$ is no less than $4$.

When $c<1$ and $\psi(c^4)$ is negligible, the optimal $\lambda$ and $\beta$ that minimize $\bar{\mathcal{D}}$ are given by
\begin{eqnarray*}
\lambda^{*}=\frac{1}{2\sigma_{\bm{w}}^2}+\frac{1}{2\sigma_{\bm{z}}^2}-\sqrt{\frac{1}{4\sigma_{\bm{z}}^4}+\frac{1}{2\sigma_{\bm{w}}^2\sigma_{\bm{z}}^2}}~\text{and}~ \beta^{*}=0.
\end{eqnarray*}

Compared with the ML estimation, the optimal $\text{MMSE}_{pb}$ denoiser reduces $\bar{\mathcal{D}}$ by a factor of
\begin{eqnarray}\label{eq:exp2_gain}
\frac{\bar{\mathcal{D}}^\text{ML}-\bar{\mathcal{D}}^{\text{MMSE}_{pb}}}{\bar{\mathcal{D}}^\text{ML}} \approx \frac{2\sigma_{\bm{z}}^2}{\sigma_{\bm{w}}^2+2\sigma_{\bm{z}}^2}.
\end{eqnarray}
\end{thm}

\begin{NewProof}
See Appendix \ref{sec:AppB}.
\end{NewProof}

\begin{figure}[t]
\centering
\subfigure{
\includegraphics[width=0.7\linewidth]{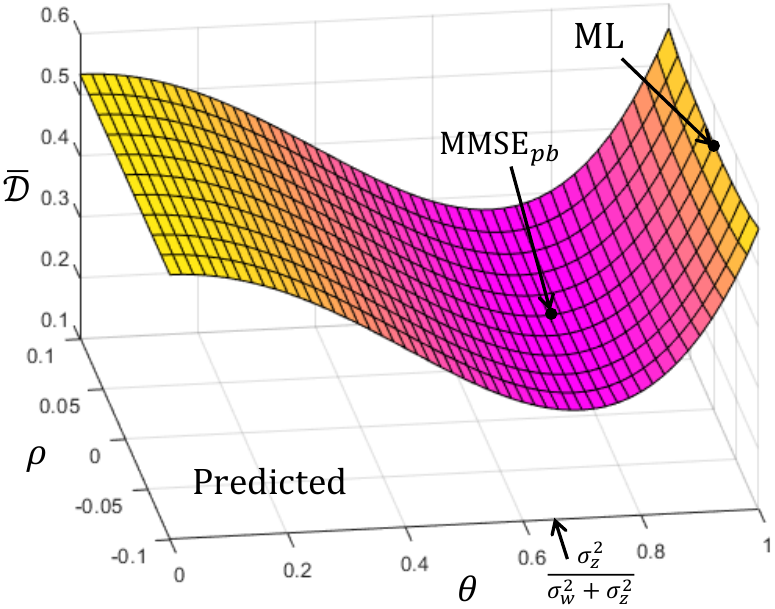}
}
\quad
\subfigure{
\includegraphics[width=0.7\linewidth]{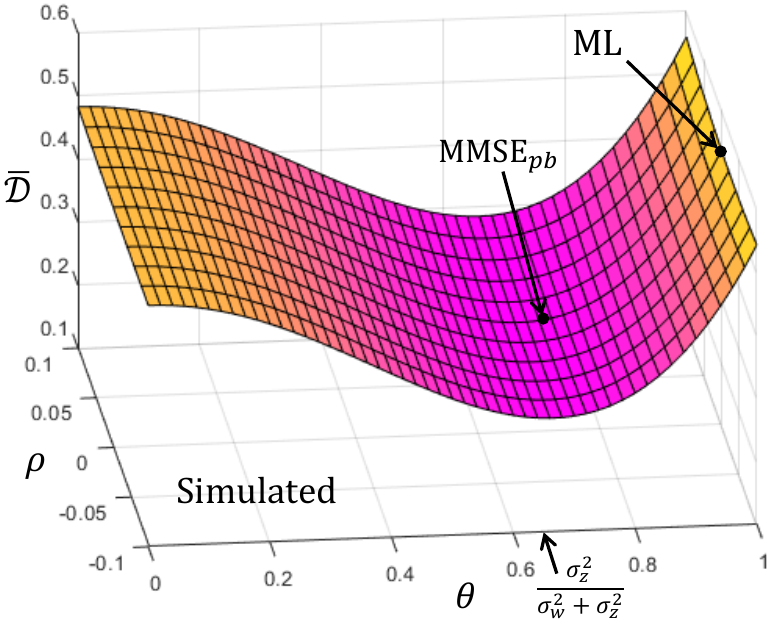}
}
\caption{The output error of a denoised NoisyNN $\bar{\mathcal{D}}$ versus $\theta$ and $\rho$. The upper figure is the analytical results in \eqref{eq:exp2_D}, where $\psi(c^4)$ is omitted; the lower figure is the simulated results of  $\bar{\mathcal{D}}$.}
\label{fig:exp2_sim12}
\end{figure}

In the following, we give an example to illustrate Theorem \ref{thm:exp2}. Specifically, we set $c = 0.4$, $\sigma^2_w=1$ and $\sigma^2_z=0.5$. In the upper half of Fig.~\ref{fig:exp2_sim12}, we plot the expected output error of the denoised neural network $\bar{\mathcal{D}}$ as a function of $\theta$ and $\rho$ according to the analytical results in \eqref{eq:exp2_D}, where the higher order terms $\psi(c^4)$ are omitted. For our $\text{MMSE}_{pb}$ denoiser, the predicted $\bar{\mathcal{D}}$ is $0.27$; while for the ML estimation, the predicted $\bar{\mathcal{D}}$ is $0.53$.
The performance gain is up to $50\%$, as per \eqref{eq:exp2_gain}.
To verify the accuracy of our predictions, we simulate $\bar{\mathcal{D}}=\mathbb{E}_{x,\bm{w},\bm{z}}(\tilde{y}-y)^2$, where $\tilde{y}$ is given by \eqref{eq:denoised2}, and plot the simulated results in the lower half of Fig.~\ref{fig:exp2_sim12}. As can be seen, the two error surfaces almost coincide with each other. The simulated $\bar{\mathcal{D}}^\text{ML}$ and $\bar{\mathcal{D}}^{\text{MMSE}_{pb}}$ are $0.25$ and $0.54$, respectively.

\begin{figure}[t]
  \centering
  \includegraphics[width=0.85\linewidth]{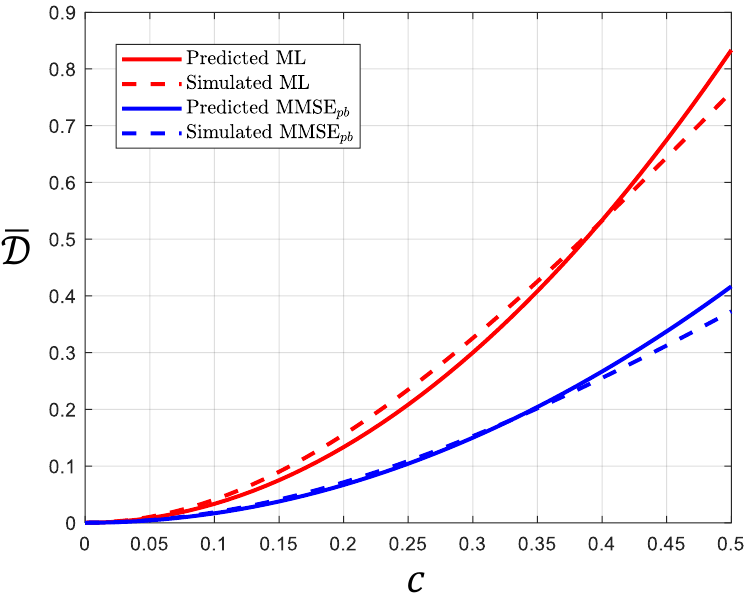}\\
  \caption{The output error $\bar{\mathcal{D}}$ versus $c$ for the ML estimation and our $\text{MMSE}_{pb}$ denoiser. The predicted performance is computed from \eqref{eq:exp2_D}, where $\psi(c^4)$ is omitted.}
\label{fig:exp2_sim3}
\end{figure}

In Theorem~\ref{thm:exp2}, we have assumed a negligible $\psi(c^4)$ to derive the optimal temperature parameters and predict the performance gain in \eqref{eq:exp2_gain}.
Next, we evaluate the impact of $c$ on the approximations, considering the same system setup as in Fig.~\ref{fig:exp2_sim12}.
In particular, we shall focus on the approximated $\bar{\mathcal{D}}$ of the ML estimation and our $\text{MMSE}_{pb}$ denoiser, benchmarked against the simulated 
$\bar{\mathcal{D}}^\text{ML}$ and 
$\bar{\mathcal{D}}^{\text{MMSE}_{pb}}$. The comparison is shown in Fig.~\ref{fig:exp2_sim3}, where we increase $c$ from $0$ to $0.5$. As can be seen, $c$ does not have to be very small. When $c\in[0,0.5]$, the differences between the predicted and simulated $\bar{\mathcal{D}}$ are marginal and the approximation is valid.

\begin{rem}
For both of the small-scale problems discussed in this section, the optimal bias compensator $\beta^*=0$ thanks to the symmetry of the neural network functions and the data statistics.
For more complex problems, such as the advanced learning tasks considered in Section~\ref{sec:5}, the symmetry no longer holds and the optimal $\beta^*\neq 0$. In those cases, analytically deriving the optimal temperature parameters is a formidable mission.
\end{rem}

\section{Numerical Experiments}\label{sec:5}
In this section, we evaluate the performance of the $\text{MMSE}_{pb}$ denoiser on advanced learning tasks with modern DNN architectures.
We consider a computer vision task and an NLP task, respectively, and implement different DNN architectures to study the performance of the denoiser.

\subsection{Learning tasks and DNNs}
For the computer vision task, we focus on the  CIFAR-10 image classification problem \cite{CIFAR10}. 
The CIFAR-10 dataset (MIT license) consists of $60,000$ $32\times 32$ colour images in $10$ classes, with $6,000$ images per class. There are $50,000$ training images and $10,000$ test images. We implement three DNNs: ResNet34, ResNet18, and ShuffleNet V2, the detailed architectures of which can be found in \cite{he2016deep} and \cite{Shufflenet}, respectively. 

For the NLP task, we consider the SST-2 sentiment classification problem \cite{SST2}.
The SST-2 dataset (GNU general public license) is a corpus with fully labeled parse trees that allows for a complete analysis of the compositional effects of sentiment in language. The corpus consists of a training dataset of $6,920$ examples, a validation dataset of $872$ examples, and a test dataset of $1,821$ examples, wherein each example is labelled as either positive or negative. We consider the state-of-the-art DNN architecture: Bidirectional Encoder Representations from Transformers (BERT) \cite{Transformer,bert}. 

The experiments were conducted on a Linux server with two CPUs (Intel Xeon E5-2643) and four GPUs (GeForce GTX 1080Ti).\footnote{Our source codes, data, and well-trained DNNs are available online at \url{https://github.com/lynshao/NoisyNN}.} 

\begin{figure*}[t]
\centering
\subfigure{
\includegraphics[width=0.37\linewidth]{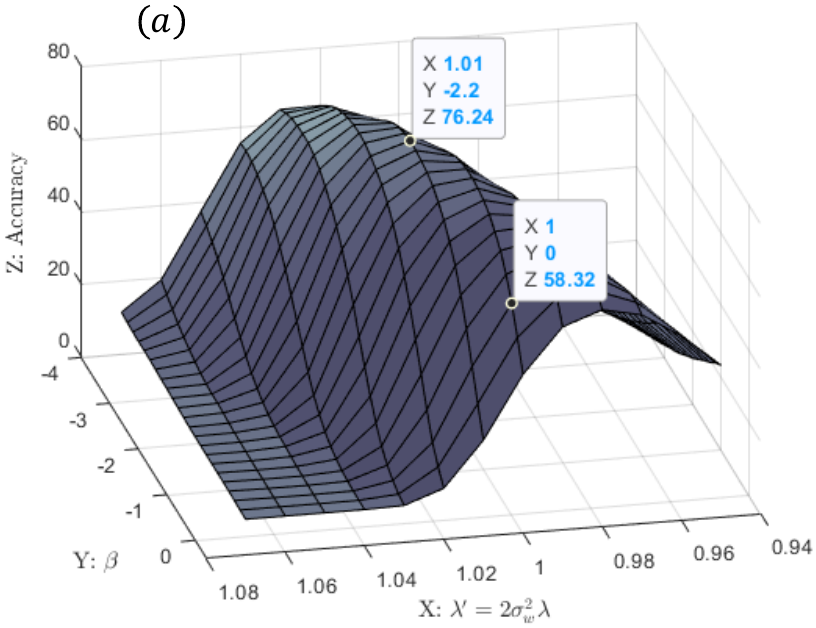}
}
\quad
\subfigure{
\includegraphics[width=0.37\linewidth]{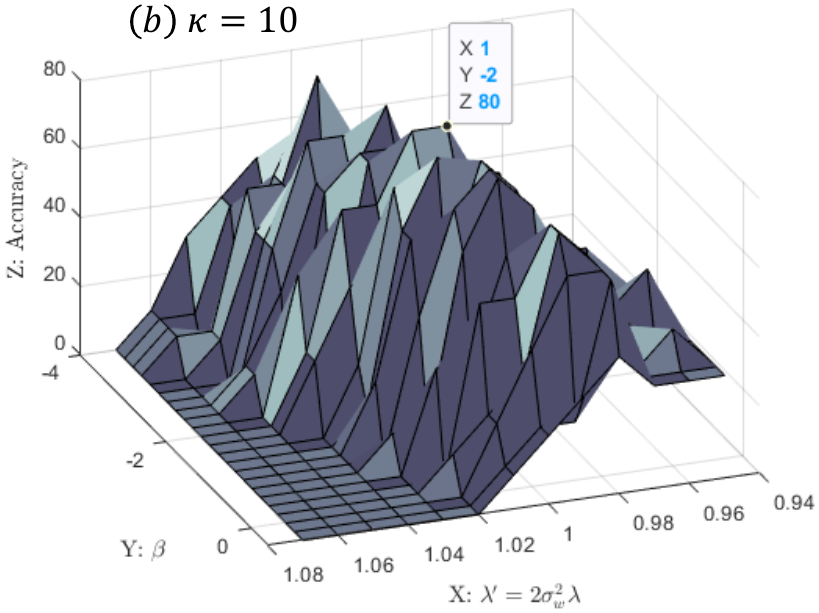}
}
\quad
\subfigure{
\includegraphics[width=0.37\linewidth]{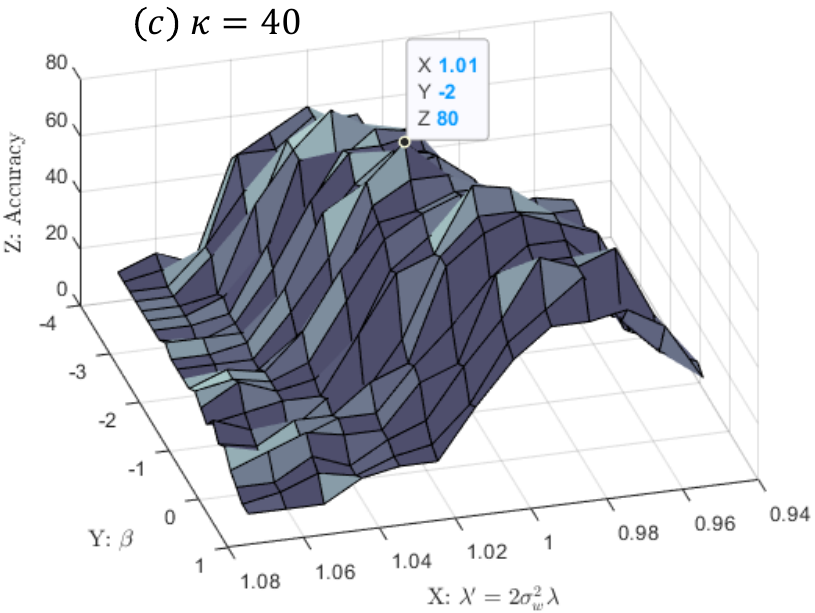}
}
\quad
\subfigure{
\includegraphics[width=0.37\linewidth]{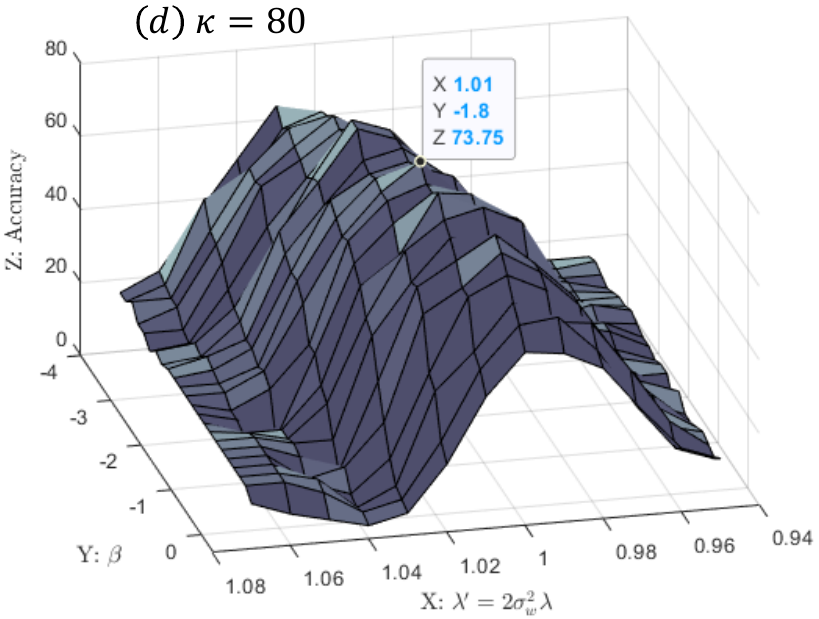}
}
\caption{Results of grid search on the test dataset (figure (a)) and validation dataset (figures (b-d)). The $x$-axis is $\lambda'=2\sigma^2_w\lambda$, the $y$-axis is $\beta$, and the $z$-axis is the test (validation) accuracy achieved by the denoised model on the test (validation) dataset. Noise is added to the well-trained ResNet34 according to a fixed WNR $\eta=-7$ dB. When $\lambda^\prime=1$ and $\beta=0$, the $\text{MMSE}_{pb}$ denoiser reduces to the ML estimator.}
\label{fig:S1}
\end{figure*}

\subsection{Noisy inference}\label{sec:VB}
In the first part, we consider the class of NoisyNNs arising from noisy inference, where noise is introduced only in the inference phase and the training phase is the standard centralized and noiseless training via backpropagation. For benchmarking purposes, we train three DNN models (a ResNet34, a ResNet18, and a ShuffleNet V2) on the CIFAR-10 dataset, and a BERT model on the SST-2 dataset in a noiseless and centralized manner. After training, the test accuracy (i.e., the prediction accuracy of the learned model on the test dataset) achieved by the four DNN models on their respective test datasets are $95.81\%$, $95.35\%$, $92.12\%$, and $92.70\%$, respectively, as shown in Table~\ref{tab:1}.

\begin{table}
  \caption{Test accuracy (with centralized and noiseless training) achieved by different DNN models on CIFAR-10 and SST-2.}
  \label{tab:1}
  \centering
  \begin{tabular}{ccccc}
    \toprule
    DNNs & \makecell[c]{ResNet34 \\(CIFAR-10)} & \makecell[c]{ResNet18 \\(CIFAR-10)}     & \makecell[c]{ShuffleNet V2\\ (CIFAR-10)} & \makecell[c]{BERT \\(SST-2)} \\
    \midrule
    \# paras       & $21.3$M        & $11.2$M        & $1.25$M  & $109.5$M  \\
    \makecell[c]{Accuracy}     & $95.81\%$ & $95.35\%$ & $92.12\%$ & $92.70\%$ \\
    \bottomrule
  \end{tabular}
\end{table}

\subsubsection{NoisyNN}
For each well-trained DNN, we add AWGN to the weights according to a given weight variance to noise power ratio (WNR) in dB, i.e., $\eta=10\log_{10}(\sigma^2_w/\sigma^2_z)$.
A caveat here is that the parameters of the batch-nor\-maliza\-tion and layer-nor\-maliza\-tion layers are set to be noise-free in the experiments \cite{Nature1,zhou1}, because they behave differently than other parameters \cite{TAMU}. These parameters are few in number. In practice, 
we can transmit/store them in a reliable manner (for example, via digital com\-munication/storage \cite{Nature1} or protect them by repetition coding \cite{Isik}).

Then, we denoise the NoisyNNs using the ML estimator and the $\text{MMSE}_{pb}$ denoiser, respectively. The denoised models will be evaluated on the test dataset to obtain test accuracy as the performance indicator of the denoiser. 

\begin{figure*}[t]
\centering
\subfigure{
\includegraphics[width=0.35\linewidth]{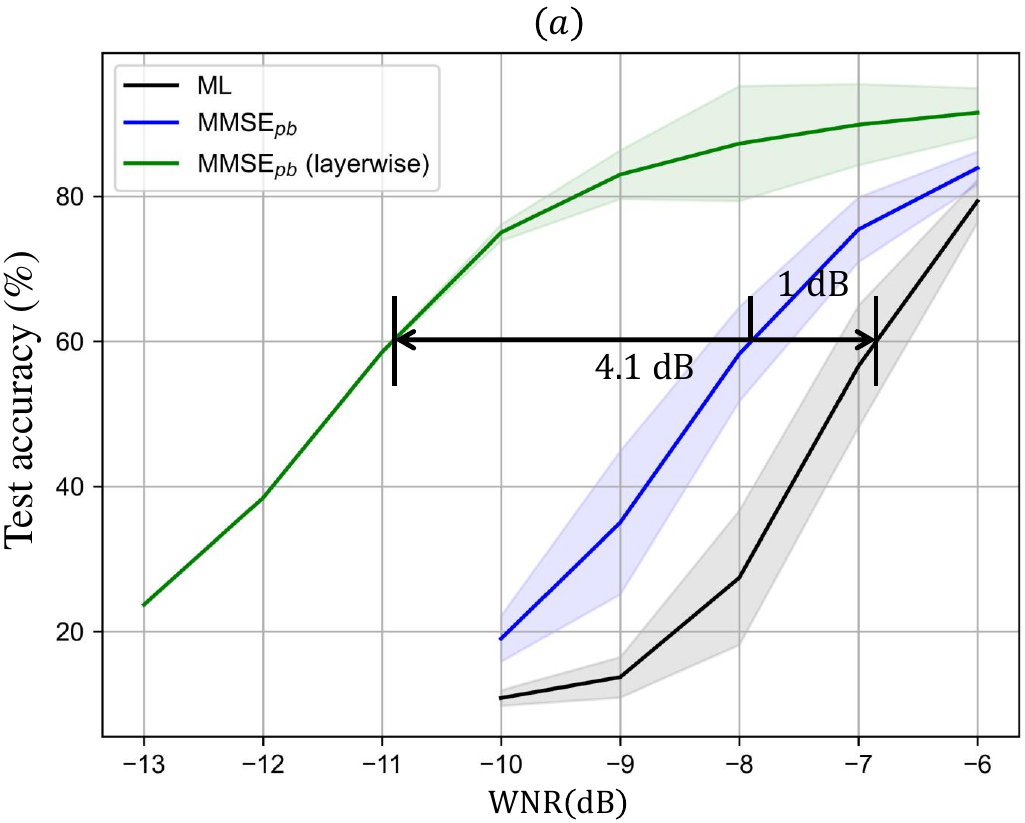}
}
\quad
\subfigure{
\includegraphics[width=0.35\linewidth]{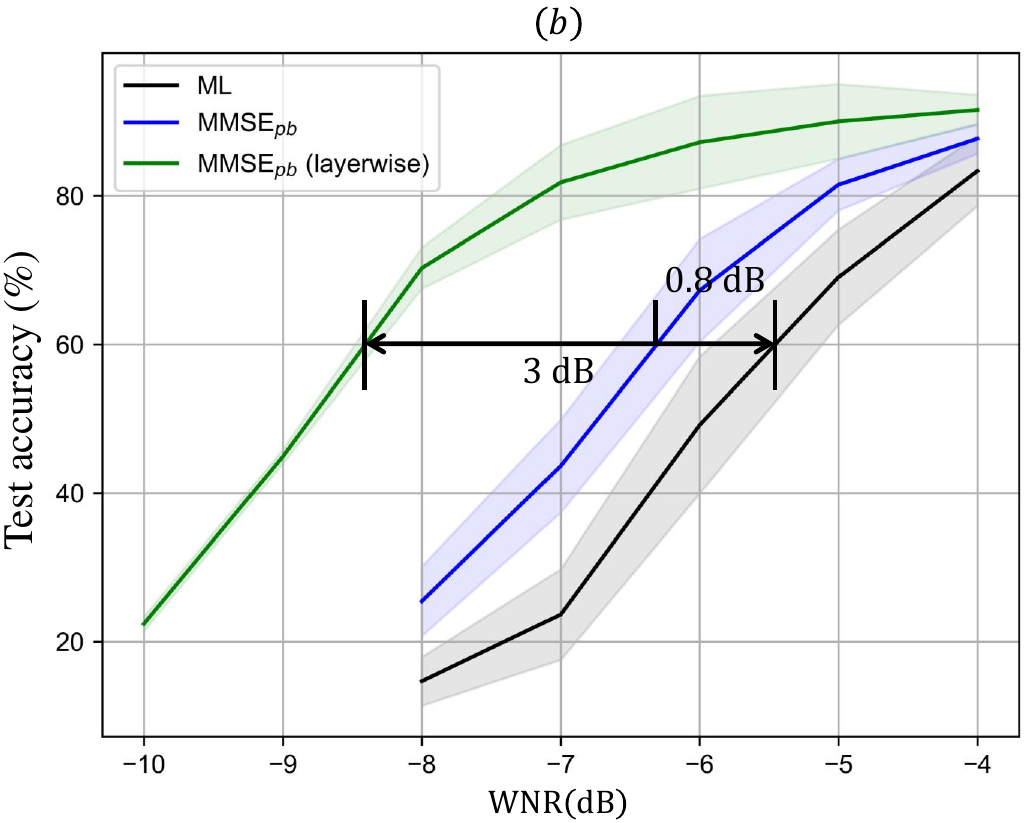}
}
\quad
\subfigure{
\includegraphics[width=0.35\linewidth]{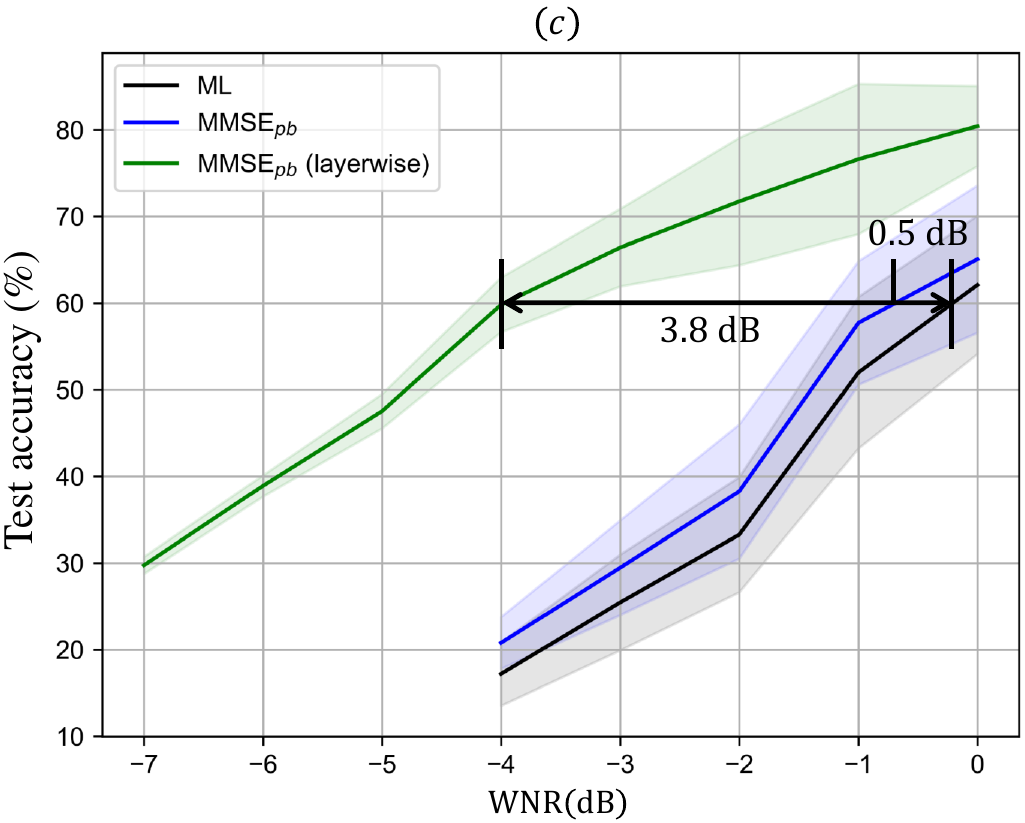}
}
\quad
\subfigure{
\includegraphics[width=0.35\linewidth]{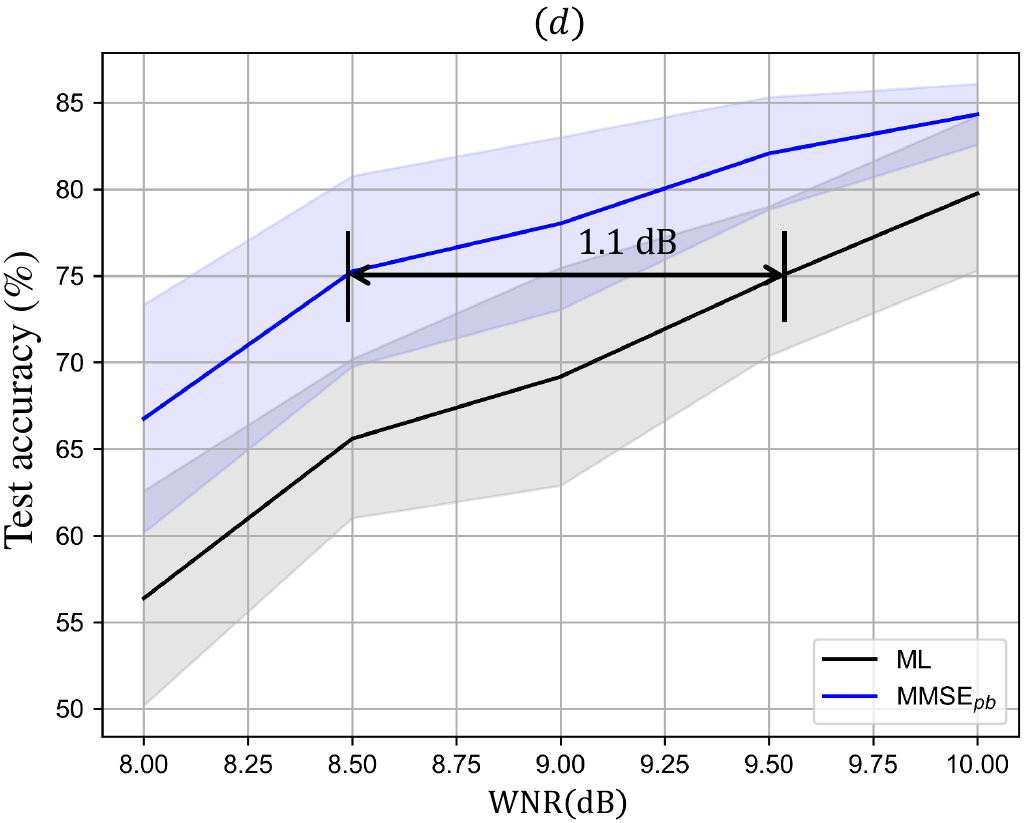}
}
\caption{The average test accuracy of the denoised models on the CIFAR-10 and SST-2 datasets with the $\text{MMSE}_{pb}$ denoiser and the ML estimator: (a) RestNet34 (CIFAR-10); (b) RestNet18 (CIFAR-10); (c) ShuffleNet V2 (CIFAR-10); (d) BERT (SST-2).}
\label{fig:S3}
\end{figure*}

\subsubsection{Determining the temperature parameters}
For the considered DNN models and learning tasks in this section, analytically deriving the optimal temperature parameters for the $\text{MMSE}_{pb}$ denoiser is no longer possible. Instead, we will identify the temperature parameters by a grid search scheme.
For convenience of notation, we define a normalized temperature parameter $\lambda^\prime=2\sigma^2_w\lambda$ to replace $\lambda$ -- our goal is to discover $\lambda^{\prime}$ and $\beta$ in the following.

Let us focus on the experiments of ResNet34 for the CIFAR-10 task.
The optimal temperature parameters $\lambda^{\prime *}$ and $\beta^*$ can be discovered by a grid search on the test dataset.
Specifically, we try various combinations of $\lambda'$ and $\beta$ to denoise the noisy ResNet34 and find the optimal $\lambda^{\prime *}$ and $\beta^*$ that yield the maximum test accuracy on the test dataset.
Fig.~\ref{fig:S1}(a) presents the results of such a grid search. As can be seen, different combinations of $\lambda'$ and $\beta$ yield different test accuracy. The optimal temperature parameters that give us the maximum test accuracy are $\lambda^{\prime *}=1.01$ and $\beta^*=-2.2$.
Note that when $\lambda^\prime=1$ and $\beta=0$, the $\text{MMSE}_{pb}$ denoiser reduces to the ML estimator. As shown in Fig.~\ref{fig:S1}(a), the gain of our $\text{MMSE}_{pb}$ denoiser over the ML estimator is up to $17.92\%$.

What Fig.~\ref{fig:S1}(a) shows is the maximum gain one can achieve from the $\text{MMSE}_{pb}$ denoiser with the optimum $\lambda^{\prime *}$ and $\beta^*$. In practice, the receiver/observer cannot perform the grid search on the test dataset due to the absence of ``labels'' of the test dataset, and hence, the optimum temperature parameters are unavailable. Nevertheless, the receiver/observer can perform the grid search on a validation dataset and use the temperature parameters that are optimal to the validation dataset (but may be suboptimal to the test dataset) to denoise the NoisyNN.
In the following, we will show that a very small validation dataset with tens of examples is sufficient to find good temperature parameters with close-to-optimal denoising performance.

Since the CIFAR-10 dataset does not have a validation dataset, we construct the validation dataset by sampling $\kappa$ examples from the training dataset. Fig.~\ref{fig:S1}(b-d) presents the results of grid search on the validation dataset, where $\kappa=10$, $40$, and $80$, respectively. Unlike Fig.~\ref{fig:S1}(a) where the $z$-axis is the test accuracy, the $z$-axis of Fig.~\ref{fig:S1}(b-d) is the validation accuracy, i.e., accuracy achieved by the denoised model on the validation dataset.

Comparing Fig.~\ref{fig:S1}(b-d) with Fig.~\ref{fig:S1}(a), the key observation is that the accuracy surface of the validation dataset is much similar to the accuracy surface of the test dataset, even when $\kappa$ is small. When $\kappa=10$, for example, the optimal temperature parameters found in the validation dataset (i.e., $\lambda^\prime=1$, $\beta=-2$) achieve a test accuracy of $72.06\%$ in Fig.~\ref{fig:S1}(a), which is very close to the optimal performance $76.24\%$. For larger $\kappa$, the achieved test accuracy is closer to the optimal, as shown in Fig.~\ref{fig:S1} (c) and (d).

\subsubsection{Experimental results}
Next, we perform extensive experiments to evaluate the performance of the $\text{MMSE}_{pb}$ denoiser benchmarked against the ML estimator, considering both CIFAR-10 and SST-2 tasks with different DNN architectures under various WNR.

The experimental results are presented in Fig.~\ref{fig:S3}, where the $x$-axis is the WNR $\eta$ and the $y$-axis is the test accuracy. In particular, to take the randomness of noise into account, we perform multiple experiments at each WNR point and plot the average test accuracy as the solid curves and the standard deviations of the achieved test accuracy as shaded areas around the average test accuracy. To find the temperature parameters, we fix the size of the validation dataset $\kappa=100$: for the CIFAR-10 task, the validation dataset is randomly sampled from the training dataset, as in Fig.~\ref{fig:S1}; for the SST-2 task, the validation dataset is randomly sampled from the default validation dataset.

As can be seen from Fig.~\ref{fig:S3}, ML estimation is in general suboptimal, our $\text{MMSE}_{pb}$ denoiser outperforms the ML estimator in all experiments.
Let us first focus on the performance of the $\text{MMSE}_{pb}$ denoiser that denoises the DNN as a whole.
For the CIFAR-10 task, with the ResNet34, ResNet18, and ShuffleNet V2 models, the average gains of the $\text{MMSE}_{pb}$ denoiser over the ML estimator are up to $1$ dB, $0.8$ dB, and $0.5$ dB, respectively, to achieve a test accuracy of $60\%$.
For the SST-2 dataset, our $\text{MMSE}_{pb}$ denoiser improves the test accuracy of the denoised BERT model by up to $1.1$ dB compared with the ML estimator to achieve a test accuracy of $75\%$.

Next, we use the $\text{MMSE}_{pb}$ denoiser to perform layer-by-layer denoising, as stated in Remark \ref{rem:layerwise}, for the three DNN models on the CIFAR-10 dataset.
As can be seen, significant performance gains are reaped.
To achieve a test accuracy of $60\%$,
the average gains of the $\text{MMSE}_{pb}$ denoiser over the ML estimator for ResNet34, ResNet18, and ShuffleNet  are up to $4.1$ dB, $3$ dB, and $3.8$ dB, respectively.\footnote{For the BERT model, larger gains are also expected with layerwise denoising. Nevertheless, due to the significant depth of the BERT architecture, search for the best temperature parameters becomes elusive.}
Another observation is that, for a given DNN, the shapes of the accuracy surface in different WNRs are similar, suggesting that we can use the same set of temperature parameters across various WNRs.

\begin{figure*}[t]
\centering
\subfigure{
\includegraphics[width=0.37\linewidth]{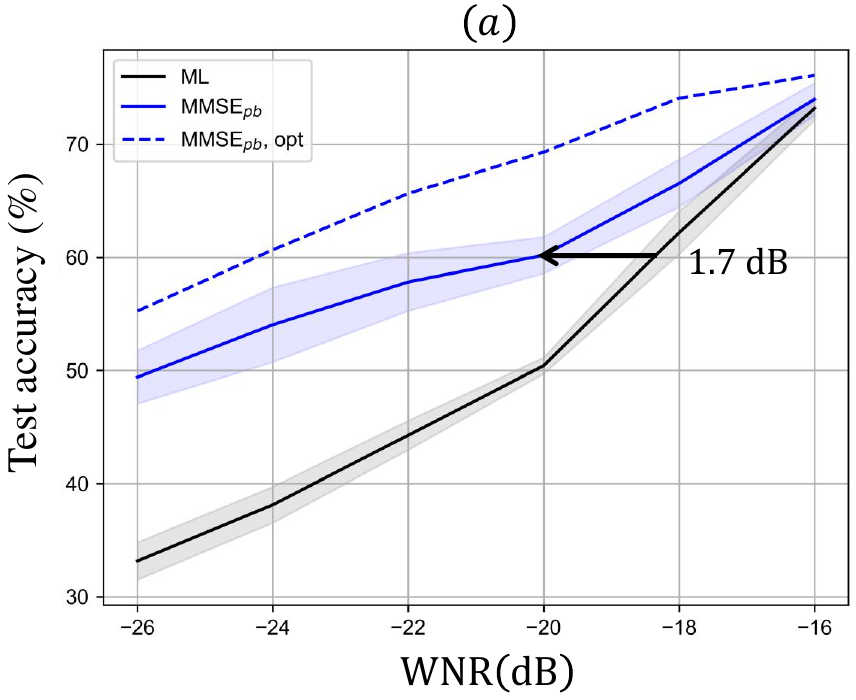}
}
\quad
\subfigure{
\includegraphics[width=0.37\linewidth]{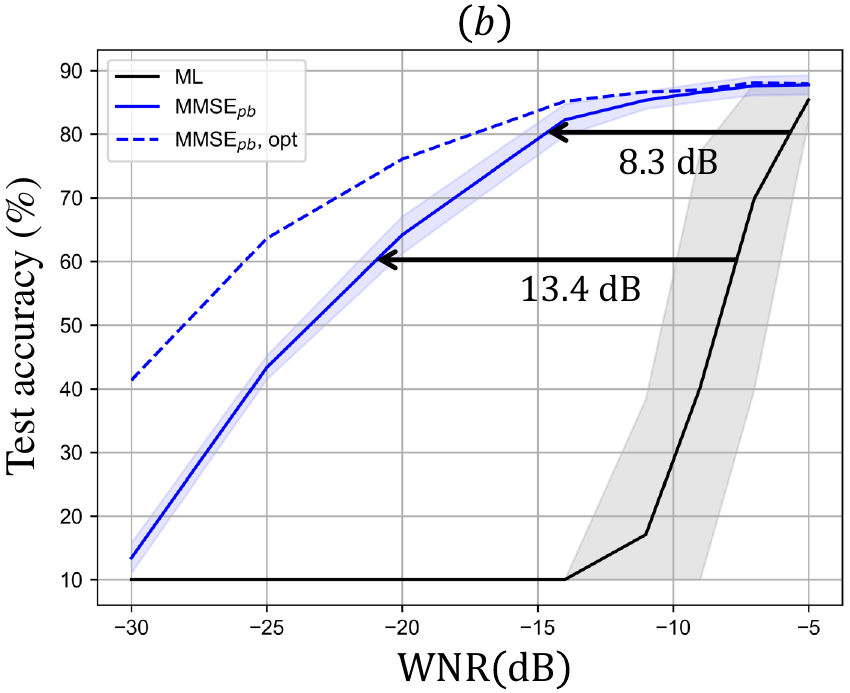}
}
\caption{The average test accuracy of the learned DNN models on CIFAR-10 dataset (with ShuffleNet V2 in (a) and ResNet18 in (b)) in FEEL with the $\text{MMSE}_{pb}$ denoiser (blue solid curves) and the ML estimator (dark solid curves). The dashed curves are the performance of the $\text{MMSE}_{pb}$ denoiser with optimized $(\lambda^{\prime*},\beta^*)$ in Section~\ref{sec:noisytraining_limit}.}
\label{fig:S6}
\end{figure*}

\subsection{Noisy training}
In the experiments of Section~\ref{sec:VB}, noise is introduced only in the inference phase on the already-trained DNNs. In this subsection, we study the performance of the denoiser in noisy training, where noise is introduced in the training phase. Specifically, we consider the application of FEEL: DNNs are trained from scratch in a distributed manner over many iterations. In each iteration, AWGN is introduced in the uplink model aggregation step and we perform $\text{MMSE}_{pb}$ denoiser at the BS to reconstruct the aggregated model.

\subsubsection{System setup}
We implement a FEEL system, wherein $20$ edge devices collaboratively train a shared model. In particular, we focus on the CIFAR-10 task and consider two lightweight DNNs, i.e., ShuffleNet V2 and ResNet18, that are more suitable for mobile deployments.

The training examples are assigned to the devices in a non-i.i.d. manner.
Recall that the CIFAR-10 dataset has a training set of $50,000$ examples and a test set of $10,000$ examples in $10$ classes. We assign non-i.i.d. training examples to the $20$ devices in the following manner:
i) first, we let each device randomly sample $2,000$ samples from the training dataset;
ii) for the remaining $10,000$ examples in the training dataset, we sort them by their labels and group them into $20$ shards of size $500$ \cite{FedAvg}. Each device is then assigned one shard.

In each iteration, $M = 4$ devices actively participating in the training. Each device trains the global model locally for $5$ epochs and transmits the model update to the BS in each iteration for aggregation. For benchmarking purposes, we train the two DNN models in a FEEL and noiseless manner (i.e., the conventional federated learning setup \cite{FedAvg}). With the non-i.i.d. assignment of training examples, the test accuracy of the ShuffleNet V2 and ResNet18 models are $83.01\%$ and $89.52\%$, respectively, when there is no noise.

\subsubsection{Experimental results}
To use the $\text{MMSE}_{pb}$ denoiser in each FEEL iteration, the BS performs a grid search on the validation dataset upon receiving the aggregated noisy model to identify the temperature parameters that yield the best validation accuracy, and then, use the temperature parameters to denoise the NoisyNNs. As in Section~\ref{sec:VB}, the validation dataset is sampled from the training dataset with $\kappa=100$ examples.

After a fixed number of FEEL iterations, we evaluate the learned model on the test dataset and present the test accuracy in Fig.~\ref{fig:S6}.
As can be seen, for the ShuffleNet V2 model, the gains of the $\text{MMSE}_{pb}$ denoiser over the ML estimator are up to $1.7$ dB to attain a test accuracy of $60\%$. When it comes to ResNet18, the gains are significantly. To achieve a test accuracy of $60\%$, the $\text{MMSE}_{pb}$ denoiser outperforms the ML estimator by $13.4$ dB. On the other hand, to achieve a test accuracy of $80\%$, the $\text{MMSE}_{pb}$ denoiser outperforms the ML estimator by $8.3$ dB.

\subsubsection{Performance limit of denoising}\label{sec:noisytraining_limit}
An important problem of our interest is the performance limit of denoising in noisy training. That is, suppose the BS has the labels of the test data (and hence, can perform the grid search on the test dataset), how can we find the optimal temperature parameters that give us the best learned models?

In noisy inference (Section~\ref{sec:VB}), NoisyNNs are generated from DNNs that are well trained -- these noiseless DNNs are supposed to achieve the maximum inference accuracy on the test dataset.
In this context, the optimal temperature parameters, and hence the optimal $\text{MMSE}_{pb}$ denoiser, are those that produce the DNN with the highest test accuracy.
Provided that the labels of the test data are known, the optimal temperature parameters can be found by a grid search on the test dataset, as what we did in Fig.~\ref{fig:S1}(a).

Similarly, in noisy training, a plausible idea to find the optimal temperature parameters in each iteration is to substitute the test dataset for the validation dataset. In other words, to denoise the aggregated model, we use the temperature parameters that yield the best test accuracy as opposed to the best validation accuracy.
This scheme, however, leads to nearly the same performance as the solid blue curves in Fig.~\ref{fig:S6}, meaning that using the validation dataset is as good as using the test dataset to identify the temperature parameters in each iteration.

An immediate question is: is this performance the limit of denoising? The answer is no.
Unlike noisy inference, the NoisyNNs in noisy training originate from half-trained DNNs -- these DNNs can perform badly on the test dataset even without noise.
In each FEEL iteration, a denoiser with temperature parameters optimized on the test dataset greedily pushes the BS to denoise the NoisyNN in a way that the reconstructed DNN performs best on the test dataset.
This is optimal for one-shot denoising, but can be suboptimal for future training/denoising and prevent the denoiser from reaping larger gains.
In this light, determining the optimal temperature parameters in successive training iterations is a stochastic planning problem and can be solved by dynamic programming.
Here, instead of an in-depth discussion, we give a simple scheme  to show that better temperature parameters is possible for our $\text{MMSE}_{pb}$ denoiser.
Specifically, as opposed to iter\-ati\-on-by-it\-era\-tion tempe\-rat\-ure-par\-ame\-ter optimization, we fix $\lambda^\prime$ and $\beta$ over the course of training and find the optimal $(\lambda^{\prime*},\beta^*)$ that produces the best test accuracy at the end of training (instead of each iteration). With this scheme, the optimized $(\lambda^{\prime*},\beta^*)$ yields better denoising performance, as depicted in Fig.~\ref{fig:S6} (the dashed curves).



\section{Conclusion}\label{sec:6}
NoisyNN is a class of DNNs whose weights are contaminated by noise. This paper puts forth a new denoising approach to reconstruct DNNs from their noisy observations or manifestations by exploiting the statistical characteristics of DNN weights.
Unlike the widely-used ML estimator that maximizes the likelihood function of the estimated DNN weights,
this work shows that focusing on minimizing the raw errors of  the DNN weights  does not align with superior DNN functional performance, because the weight errors affect the DNN performance in a nonlinear manner due to the nonlinear activation functions within DNNs.
With a goal to maximize the inference accuracy of the reconstructed model, our formulation leads to a simple and efficient $\text{MMSE}_{pb}$ denoiser.
Our approach works for NoisyNNs arising from both noisy training and noisy inference, and does not require any retraining of neural networks.
The superior performance of our denoiser over the ML estimator has been verified by both analytical and experimental results.

\appendices

\section{Proof of Theorem \ref{thm:exp1}}\label{sec:AppA}
For the quadratic neural network function in Definition \ref{defi:quadratic}, the squared output error can be written as
\begin{eqnarray*}
\mathcal{D}
\hspace{-0.25cm}&=&\hspace{-0.25cm}
\big[\widetilde{y}(\bm{x,w,z},\theta,\rho)-y(\bm{x,w})\big]^2 \\
\hspace{-0.25cm}&=&\hspace{-0.25cm}
\Big\{(\bm{x}-c\bm{e})^\top\big[ (\theta\bm{W}+\theta\bm{Z}+\rho\bm{I}_{d\times d})^2-\bm{W}^2\big] (\bm{x}-c\bm{e})\Big\}^2 \\
\hspace{-0.25cm}&\triangleq &\hspace{-0.25cm}
\Big\{(\bm{x}-c\bm{e})^\top\widetilde{\bm{W}} (\bm{x}-c\bm{e})\Big\}^2.
\end{eqnarray*}
In particular, $\widetilde{\bm{W}}$ is diagonal, the $i$-th diagonal element of which is given by
\begin{equation*}
\widetilde{w_i}=(\theta^2-1)w_i^2+2\theta^2w_iz_i+2\theta\rho w_i+\theta^2z_i^2+2\theta\rho z_i+\rho^2.
\end{equation*}

Thus, $\mathcal{D}$ can be simplified as
\begin{equation}
\mathcal{D}=\Big\{(\bm{x}-c\bm{e})^\top\widetilde{\bm{W}} (\bm{x}-c\bm{e})\Big\}^2=
\left[\sum_{i=1}^d \widetilde{w_i}(x_i-c)^2 \right]^2.
\end{equation}

We first take the expectation of $\mathcal{D}$ over $\bm{x}$. Since the elements of $\bm{x}$ are i.i.d. with $x_i \sim \mathcal{U}(-1,1)$, we have
\begin{equation*}
\mathbb{E}_{\bm{x}}\mathcal{D}=C_1\sum_{i=1}^d\widetilde{w}_i^2+C_2\sum_{i=1,j\neq i}^d\widetilde{w}_i\widetilde{w}_j,
\end{equation*}\\
where $C_1\triangleq c^4+2c^2+1/5>0$ and $C_2\triangleq(c^2-1/3)^2\geq 0$.

Next, we take the expectation of $\mathbb{E}_{\bm{x}}\mathcal{D}$ over $\bm{w}$ and $\bm{z}$. Since the elements of $\bm{w}$ and $\bm{z}$ are i.i.d., and $w_i \sim \mathcal{N}(0,\sigma_{{w}}^2)$, $z_i \sim \mathcal{N}(0,\sigma_{z}^2)$, we have
\begin{eqnarray}
\bar{\mathcal{D}} &&\hspace{-0.55cm} = \mathbb{E}_{\bm{x,w,z}}\mathcal{D}
= C_1^{\prime}\big[\theta^4(\sigma_{\bm{w}}^2+\sigma_{\bm{z}}^2)^2-2\theta^2\sigma_{\bm{w}}^4 + \\
&&\hspace{-0.5cm} 2\theta^2\rho^2(\sigma_{\bm{w}}^2+\sigma_{\bm{z}}^2)+\sigma_{\bm{w}}^4\big]
+ C_2^{\prime}\left(\rho^4-2\theta^2\sigma_{\bm{w}}^2\sigma_{\bm{z}}^2-2\sigma_{\bm{w}}^2\rho^2\right),\nonumber
\end{eqnarray}
where $C_1^{\prime}\triangleq 3C_1d+C_2d(d-1)$ and $C_2^{\prime}\triangleq C_1d+C_2d(d-1)$. Note that both $C_1^{\prime}$ and $C_2^{\prime}$ are positive since $C_1>0$, $C_2\geq0$, and $d\geq1$.

Notice that $\bar{\mathcal{D}}$ is an even function of both $\theta$ and $\rho$. We can focus on the region where $\theta\geq0$ and $\rho\geq0$, because $\bar{\mathcal{D}}$ is symmetric with respective to (w.r.t.) $\theta=0$ and $\rho=0$.

In the region $\begin{Bmatrix}\theta\geq0,\rho\geq0\end{Bmatrix}$, we examine the monotonicity of $\bar{\mathcal{D}}$ by setting its first derivative w.r.t. $\theta$ and $\rho$ to 0, giving,
\begin{eqnarray*}
\frac{\partial \bar{\mathcal{D}}}{\partial \theta}
\hspace{-0.2cm}& \propto &\hspace{-0.2cm} \theta\big[C_1^{\prime}(\sigma_{\bm{w}}^2+\sigma_{\bm{z}}^2)^2\theta^2+C_1^{\prime}(\sigma_{\bm{w}}^2+\sigma_{\bm{z}}^2)\rho^2\\
&&\hspace{2.1cm}-C_2^{\prime}\sigma_{\bm{w}}^2\sigma_{\bm{z}}^2-C_1^{\prime}\sigma_{\bm{w}}^4\big]=0 \\
\frac{\partial \bar{\mathcal{D}}}{\partial \rho}
\hspace{-0.2cm}& \propto &\hspace{-0.2cm} \rho[C_1^{\prime}(\sigma_{\bm{w}}^2+\sigma_{\bm{z}}^2)^2\theta^2-C_2^{\prime}\sigma_{\bm{w}}^2+C_2^{\prime}\rho^2]=0
\end{eqnarray*}

Thus, $\bar{\mathcal{D}}$ has four critical points in $\begin{Bmatrix}\theta\geq0,\rho\geq0\end{Bmatrix}$:
\begin{eqnarray*}
&&\hspace{-0.5cm} \text{P1}: \theta=0,~\rho=0; \\
&&\hspace{-0.5cm} \text{P2}: \theta=0,~\rho=\sigma_{\bm{w}}; \\
&&\hspace{-0.5cm} \text{P3}: \theta=\frac{\sigma^2_w}{\sigma^2_w+\sigma^2_z}\sqrt{1+\frac{C^\prime_2\sigma^2_z}{C^\prime_1\sigma^2_w}},~\rho=0; \\
&&\hspace{-0.5cm} \text{P4}: \theta=\sqrt{\frac{C_2^{\prime}}{C_1^{\prime}}}\frac{\sigma_{\bm{w}}\sigma_{\bm{z}}}{\sigma_{\bm{w}}^2+\sigma_{\bm{z}}^2},~\rho=\frac{\sigma_{\bm{w}}^2}{\sqrt{\sigma_{\bm{w}}^2+\sigma_{\bm{z}}^2}}.
\end{eqnarray*}

Then we derive the Hessian of $\bar{\mathcal{D}}$ to determine if the above critical points are local minima, maxima or saddle points. The Hessian matrix of $\bar{\mathcal{D}}$ is defined as
\begin{eqnarray}
H \triangleq
\begin{bmatrix}
\frac{\partial^2 \bar{\mathcal{D}}}{\partial \theta^2} & \frac{\partial^2 \bar{\mathcal{D}}}{\partial \theta \partial \rho} \\
\frac{\partial^2 \bar{\mathcal{D}}}{\partial \theta \partial \rho} & \frac{\partial^2 \bar{\mathcal{D}}}{\partial \rho^2}
\end{bmatrix},
\end{eqnarray}
where
\begin{eqnarray*}
\frac{\partial^2 \bar{\mathcal{D}}}{\partial \theta^2}
\hspace{-0.2cm}& = &\hspace{-0.2cm} 4\big[3C_1^{\prime}(\sigma_{\bm{w}}^2+\sigma_{\bm{z}}^2)^2\theta^2-C_2^{\prime}\sigma_{\bm{w}}^2\sigma_{\bm{z}}^2-C_1^{\prime}\sigma_{\bm{w}}^4 \\
&&\hspace{0.1cm}+C_1^{\prime}(\sigma_{\bm{w}}^2+\sigma_{\bm{z}}^2)\rho^2\big];  \\
\frac{\partial^2 \bar{\mathcal{D}}}{\partial \rho^2}
\hspace{-0.2cm}& = &\hspace{-0.2cm}
4\big[C_1^{\prime}(\sigma_{\bm{w}}^2+\sigma_{\bm{z}}^2)^2\theta^2-C_2^{\prime}\sigma_{\bm{w}}^2+3C_2^{\prime}\rho^2\big]; \\
\frac{\partial^2 \bar{\mathcal{D}}}{\partial \theta \partial \rho}
\hspace{-0.2cm}& = &\hspace{-0.2cm}
8C_1^{\prime}\big(\sigma_{\bm{w}}^2+\sigma_{\bm{z}}^2\big)\theta\rho.
\end{eqnarray*}

Next, we examine the positive definiteness of the Hessian at the critical points $\text{P1}\sim \text{P4}$.

For P1, $\forall\bm{u}=[u_1,u_2]^\top\neq \bm{0}$, we have
\begin{equation*}
\bm{u}^T\bm{H}\bm{u}=-4(C_2^{\prime}\sigma_{\bm{w}}^2\sigma_{\bm{z}}^2+C_1^{\prime}\sigma_{\bm{w}}^4)u_1^2-4C_2^{\prime}\sigma_{\bm{w}}^2u_2^2<0.
\end{equation*}

For P2, $\forall\bm{u}=[u_1,u_2]^\top\neq \bm{0}$, we have
\begin{equation*}
\bm{u}^T\bm{H}\bm{u}=8dC_1\sigma_{\bm{w}}^2\sigma_{\bm{z}}^2u_1^2+8C_2^{\prime}\sigma_{\bm{w}}^2u_2^2>0.
\end{equation*}

For P3, $\forall\bm{u}=[u_1,u_2]^\top\neq \bm{0}$, we have
\begin{equation*}
\bm{u}^T\bm{H}\bm{u}=(8C_2^{\prime}\sigma_{\bm{w}}^2\sigma_{\bm{z}}^2+8C_1^{\prime}\sigma_{\bm{w}}^4)u_1^2+\frac{8dC_1\sigma_{\bm{w}}^4}{(\sigma_{\bm{w}}^2+\sigma_{\bm{z}}^2)}u_2^2>0.
\end{equation*}

For P4, $\forall\bm{u}=[u_1,u_2]^\top\neq \bm{0}$, we have
\begin{eqnarray*}
&&\hspace{-0.5cm}\bm{u}^T\bm{H}\bm{u} =
8C_2^{\prime}\sigma_{\bm{w}}^2\sigma_{\bm{z}}^2u_1^2+8C_2^{\prime}\frac{\sigma_{\bm{w}}^4}{(\sigma_{\bm{w}}^2+\sigma_{\bm{z}}^2)}u_2^2+ \\
&&\hspace{1.1cm} 16\sqrt{\frac{{C_2^{\prime}}^2+2C_1C_2^{\prime}}{\sigma_{\bm{w}}^2+\sigma_{\bm{z}}^2}}\sigma_{\bm{w}}^3\sigma_{\bm{z}}u_1u_2 \\
&&\hspace{-0.5cm} > 8C_2^{\prime}\begin{bmatrix}\sigma_{\bm{w}}^2\sigma_{\bm{z}}^2u_1^2+\frac{\sigma_{\bm{w}}^4}{(\sigma_{\bm{w}}^2+\sigma_{\bm{z}}^2)}u_2^2+\frac{2}{\sqrt{\sigma_{\bm{w}}^2+\sigma_{\bm{z}}^2}}\sigma_{\bm{w}}^3\sigma_{\bm{z}}u_1u_2\end{bmatrix}  \\
&&\hspace{-0.5cm} = 8C_2^{\prime}\begin{pmatrix}\sigma_{\bm{w}}\sigma_{\bm{z}}u_1+\frac{\sigma_{\bm{w}}^2}{\sqrt{\sigma_{\bm{w}}^2+\sigma_{\bm{z}}^2}}u_2\end{pmatrix}^2\geq 0.
\end{eqnarray*}

As a result, P1 is a local maximum, and P2, P3, P4 are local minima.

Notice that an additional constraint in \eqref{eq:MMSEpb} is $0\leq\lambda<\frac{1}{2\sigma^2_w}+\frac{1}{2\sigma^2_z}$. Therefore, a feasible $\theta$ must satisfy $\frac{\sigma^2_w}{\sigma^2_w+\sigma^2_z}\leq\theta<\infty$. This means that the optimal $\theta$ and $\rho$ that minimizes $\bar{\mathcal{D}}$ should be chosen between P3 and P4.

At P3 and P4, we have
\begin{eqnarray*}
\bar{\mathcal{D}}(\text{P3})
\hspace{-0.2cm}& = &\hspace{-0.2cm}
(C_1^{\prime}-C_2^{\prime})\frac{(C_1^{\prime}+C_2^{\prime})\sigma_{\bm{z}}^2+2C_1^{\prime}\sigma_{\bm{w}}^2}{C_1^{\prime}(\sigma_{\bm{w}}^2+\sigma_{\bm{z}}^2)^2}\sigma_{\bm{w}}^4\sigma_{\bm{z}}^2, \\
\bar{\mathcal{D}}(\text{P4})
\hspace{-0.2cm}& = &\hspace{-0.2cm}
(C_1^{\prime}-C_2^{\prime})\begin{pmatrix}1+\frac{C_2^{\prime}\sigma_{\bm{z}}^4}{C_1^{\prime}(\sigma_{\bm{w}}^2+\sigma_{\bm{z}}^2)^2}\end{pmatrix}\sigma_{\bm{w}}^4.
\end{eqnarray*}

It is easy to show that
\begin{equation}
\bar{\mathcal{D}}(\text{P4}) - \bar{\mathcal{D}}(\text{P3})
= 2C_1 d\frac{\sigma^8_w}{(\sigma^2_w+\sigma^2_z)^2} > 0,
\end{equation}
and hence, P3 is the global minimum, in which case
\begin{equation*}
\lambda=\frac{1}{2\sigma^2_w}+\frac{1}{2\sigma^2_z}-\sqrt{\frac{C^\prime_1\sigma^2_w}{C^\prime_1\sigma^2_w+C^\prime_2\sigma^2_z}}\frac{\sigma^2_w+\sigma^2_z}{2\sigma^2_w\sigma^2_z},~\beta=0.
\end{equation*}

It is worth noting that, by symmetry, another global minimum is
\begin{equation*}
\text{P3}^\prime: \theta=-\frac{\sigma^2_w}{\sigma^2_w+\sigma^2_z}\sqrt{1+\frac{C^\prime_2\sigma^2_z}{C^\prime_1\sigma^2_w}},~\rho=0; 
\end{equation*}
but only P3 satisfies the constraint that $\theta\geq \frac{\sigma^2_w}{\sigma^2_w+\sigma^2_z}$.

Finally, we evaluate the performance gain of the optimal $\text{MMSE}_{pb}$ denoiser over the ML estimation.
Let $\lambda=\frac{1}{2\sigma^2_w}$ and $\beta=0$, we have
\begin{equation}
\bar{\mathcal{D}}^{\text{ML}}=C^\prime_1\sigma^4_z + 4C_1 d\sigma^2_w\sigma^2_z.
\end{equation}
It is easy to verify that \eqref{eq:factor} is true.

\section{Proof of Theorem \ref{thm:exp2}}\label{sec:AppB}
Considering the denoised NoisyNN in \eqref{eq:denoised2}, the expected output error can be written as
\begin{eqnarray*}
\bar{D}(\theta,\rho)
\hspace{-0.2cm}&=&\hspace{-0.2cm} 
\mathbb{E}_{x,\bm{w},\bm{z}}\mathcal{D} \\
\hspace{-0.2cm}&=&\hspace{-0.2cm} \mathbb{E}_{x,\bm{w},\bm{z}}[\tilde{y}(x,\bm{w},\bm{z},\theta,\rho)-y(x,\bm{w})]^2.
\end{eqnarray*}

As a first step, we take expectation of $\mathcal{D}$ over $\bm{z}$, yielding
\begin{equation}\label{eq:B10}
\mathbb{E}_{\bm{z}}\mathcal{D} = \mathbb{E}_{\bm{z}} \tilde{y}^2 - 2y \mathbb{E}_{\bm{z}} \tilde{y} + y^2.
\end{equation}

In Lemma \ref{lem:app}, we have approximated $\mathbb{E}_{\bm{z}}\tilde{y}$ by \eqref{eq:lemma1}. Using the same approach, we can approximate $\mathbb{E}_{\bm{z}} \tilde{y}^2$ as
\begin{eqnarray}
\mathbb{E}_{\bm{z}} \tilde{y}^2
\hspace{-0.2cm}&\approx&\hspace{-0.2cm}  \tilde{y_0}^2 + \mathbb{E}_{\bm{z}} \frac{1}{2} \bm{z}^\top \frac{\partial^2 \tilde{y_0}^2}{\partial \bm{w}^2} \bm{z} \nonumber\\
\hspace{-0.2cm}&=&\hspace{-0.2cm} \tilde{y_0}^2 + \frac{\sigma_{\bm{z}}^2}{2} \left( \sum_{i=1}^N \frac{\partial^2 \tilde{y_0}^2}{\partial v_i^2} + \sum_{i=1}^N \frac{\partial^2 \tilde{y_0}^2}{\partial u_i^2} \right),
\end{eqnarray}
where
\begin{eqnarray*}
\frac{\partial^2 \tilde{y_0}^2}{\partial v_i^2}
\hspace{-0.2cm}&=&\hspace{-0.2cm}
2 \left( \frac{\partial \tilde{y_0}}{\partial v_i} \right)^2 + 2\tilde{y_0}\frac{\partial^2 \tilde{y_0}}{\partial v_i^2} = 2\theta^2 g_i^2, \\
\frac{\partial^2 \tilde{y_0}^2}{\partial u_i^2}
\hspace{-0.2cm}&=&\hspace{-0.2cm}
2(\theta v_i + \rho)^2 \theta^2 x^2 {g_i^{\prime}}^2 + 4\tilde{y_0}(\theta v_i + \rho) \theta^2 x^2 g_i^{\prime \prime}.
\end{eqnarray*}

Thus,
\begin{eqnarray}\label{eq:B11}
\mathbb{E}_{\bm{z}} \tilde{y}^2
\hspace{-0.2cm}&\approx&\hspace{-0.2cm} \tilde{y_0}^2 + \sigma_{\bm{z}}^2 \sum_{i=1}^N \Big[\theta^2 g_i^2 + (\theta v_i + \rho)^2 \theta^2 x^2 {g_i^{\prime}}^2 + \nonumber\\
&&\hspace{1.8cm}2\tilde{y_0}(\theta v_i + \rho) \theta^2 x^2 g_i^{\prime \prime}\Big].
\end{eqnarray}

Substituting \eqref{eq:lemma1} and \eqref{eq:B11} into \eqref{eq:B10} gives us
\begin{eqnarray}
 \mathbb{E}_{\bm{z}}\mathcal{D}
 \hspace{-0.2cm}&\approx&\hspace{-0.2cm}
 \underbrace{(\tilde{y_0} - y)^2}_{A} + \underbrace{\sigma_{\bm{z}}^2 \theta^2 \sum_{i=1}^N \left[g_i^2 + (\theta v_i + \rho)^2 x^2 {g_i^{\prime}}^2 \right]}_{B} + \nonumber\\
 &&\hspace{-0.2cm}\underbrace{2 \sigma_{\bm{z}}^2 (\tilde{y_0}-y) \theta^2 x^2 \sum_{i=1}^N (\theta v_i + \rho) g_i^{\prime \prime}}_{C}.
\end{eqnarray}

In the following, we derive $A$, $B$, and $C$, respectively, and take expectation over the input $x$ and the weight vector $\bm{w}$.

Let us start from $A$.
\begin{eqnarray}\label{eq:B_A}
&&\hspace{-0.5cm} A = (\tilde{y_0} - y)^2 \\
&&\hspace{-0.5cm} = \underbrace{\sum_{i,j} (\theta v_i \!+\! \rho)\tanh(\theta u_i x \!+\! \rho x)(\theta v_j \!+\! \rho)\tanh(\theta u_j x \!+\! \rho x)}_{A_1} \nonumber\\
&&\hspace{-0.1cm} - 2 \underbrace{\sum_{i,j}v_i \tanh(u_i x)(\theta v_j + \rho)\tanh(\theta u_j x + \rho x)}_{A_2}\nonumber\\
&&\hspace{-0.1cm} + \underbrace{\sum_{i,j}\tanh(u_i x) v_j \tanh(u_j x)}_{A_3},\nonumber
\end{eqnarray}
where $i=1,2,...,N$ and $j=1,2,...,N$.

First, $A_1$ can be further written as
\begin{eqnarray*}
&&\hspace{-0.6cm} A_1 = \underbrace{\sum_{i=1}^N (\theta v_i + \rho)^2 \tanh^2(\theta u_i x + \rho x)}_{A_{11}}  + \\
&&\hspace{-0.6cm} \underbrace{\sum_{{i,j,i\neq j}}(\theta v_i \!+\! \rho)\tanh(\theta u_i x \!+\! \rho x)(\theta v_j \!+\! \rho)\tanh(\theta u_j x \!+\! \rho x)}_{A_{12}}.
\end{eqnarray*}

The $\tanh$ function can be approximated by $\tanh(\alpha) = \alpha - \frac{1}{3} \alpha^3 + \frac{2}{15} \alpha^5 - ... \approx \alpha - \frac{1}{3} \alpha^3$. Therefore,
\begin{eqnarray*}
A_{11}
\hspace{-0.2cm}&\approx&\hspace{-0.2cm} \sum_{i=1}^N (\theta^2 v_i^2 + 2\theta \rho v_i + \rho^2)\Big[(\theta u_i x + \rho x)^2 \\
&&\hspace{1cm} - \frac{2}{3}(\theta u_i x + \rho x)^4 + \frac{1}{9}(\theta u_i x + \rho x)^6 \Big].
\end{eqnarray*}

Next we take the expectation over $x$. Since $x\! \sim\! \mathcal{U}(-c,c)$, we have
\begin{equation*}
\mathbb{E}_x x^m = \frac{c^{m+1} - (-1)^{m+1} c^{m+1}}{(m+1)   2c} = c^m \frac{1-(-1)^{m+1}}{2(m+1)}
\end{equation*}
Thus,
\begin{equation*}
\mathbb{E}_x A_{11} \approx \frac{1}{3} \sum_{i=1}^N (\theta^2 v_i^2 + 2\theta \rho v_i + \rho^2)(\theta u_i + \rho)^2 c^2 + \psi(c^4).
\end{equation*}

Then we take the expectation over $\bm{w}=\left\{\bm{u},\bm{v}\right\}$. This gives us
\begin{equation}
\mathbb{E}_x \mathbb{E}_{\bm{w}} A_{11} \approx \frac{N}{3}(\theta^2 \sigma_{\bm{w}}^2 + \rho^2)^2 c^2 + \psi(c^4).
\end{equation}

Likewise, for $A_{12}$, we have
\begin{equation*}
\mathbb{E}_x \mathbb{E}_{\bm{w}} A_{12} \approx N(N\!-\!1)\left[ \frac{1}{3} \rho^4 c^2 \!+\! \left(\theta^2 \rho^2 \sigma_{\bm{w}}^2 \!+\! \frac{1}{3} \rho^4 \right)^2\right] \!+\! \psi(c^4).
\end{equation*}

The other two terms, $A_2$ and $A_3$ in \eqref{eq:B_A}, can be derived in the similar way. Despite the intricate derivations, the final form of $\mathbb{E}_x \mathbb{E}_{\bm{w}} A_2$ and $\mathbb{E}_x \mathbb{E}_{\bm{w}} A_3$ are quite neat:
\begin{eqnarray*}
\mathbb{E}_x \mathbb{E}_{\bm{w}} A_2
\hspace{-0.2cm}&\approx&\hspace{-0.2cm}
\frac{N}{3} \theta^2 c^2 \sigma_{\bm{w}}^4 + \psi(c^4), \\
\mathbb{E}_x \mathbb{E}_{\bm{w}} A_3
\hspace{-0.2cm}&\approx&\hspace{-0.2cm}
\frac{N}{3}\sigma_{\bm{w}}^4 c^2 + \psi(c^4).
\end{eqnarray*}

Substituting $\mathbb{E}_x \mathbb{E}_{\bm{w}} A_1$, $\mathbb{E}_x \mathbb{E}_{\bm{w}} A_2$ and $\mathbb{E}_x \mathbb{E}_{\bm{w}} A_3$ into  \eqref{eq:B_A}, we have
\begin{eqnarray*}
&&\hspace{-0.5cm} \mathbb{E}_x \mathbb{E}_{\bm{w}} A \approx  \frac{N}{3}(\theta^2 \sigma_{\bm{w}}^2 + \rho^2)^2 c^2 + N(N-1)\Big[ \frac{1}{3} \rho^4 c^2 + \\
&&\hspace{-0.1cm} \left(\theta^2 \rho^2 \sigma_{\bm{w}}^2 + \frac{1}{3} \rho^4 \right)^2\Big] - \frac{2N}{3} \theta^2 c^2 \sigma_{\bm{w}}^4 + \frac{N}{3}\sigma_{\bm{w}}^4 c^2 + \psi(c^4).
\end{eqnarray*}

Then, for $B$, we first take the expectation over $\bm{v}$, yielding
\begin{eqnarray*}
&&\hspace{-0.5cm} \mathbb{E}_{\bm{v}}B = \mathbb{E}_{\bm{v}}\sigma_{\bm{z}}^2 \theta^2 \sum_{i=1}^N \left[g_i^2 + (\theta v_i + \rho)^2 x^2 {g_i^{\prime}}^2 \right] \\
&&\hspace{-0.5cm} = \sigma_{\bm{z}}^2 \theta^2\sum_{i=1}^N \big[\tanh^2(\theta u_i x + \rho x) + (\theta^2 \sigma_{\bm{w}}^2 + \rho^2)x^2\times \\
&&\hspace{3cm} (1 - \tanh^2(\theta u_i x + \rho x))^2 \big] \\
&&\hspace{-0.5cm} \overset{(a)}{\approx} \sigma_{\bm{z}}^2 \theta^2 \sum_{i=1}^N \tilde{\bm{x}}^\top
\begin{bmatrix}
(\theta u_i + \rho)^2 + (\theta^2 \sigma_{\bm{w}}^2 + \rho^2)\\
-\frac{2}{3} (\theta u_i \!+\! \rho)^4 \!-\! 2 (\theta u_i \!+\! \rho)^2 (\theta^2 \sigma_{\bm{w}}^2 \!+\! \rho^2)\\
\frac{1}{9} (\theta u_i \!+\! \rho)^6 + \frac{7}{3}(\theta u_i \!+\! \rho)^4 (\theta^2 \sigma_{\bm{w}}^2 \!+\! \rho^2)\\
-\frac{14}{9} (\theta u_i + \rho)^6 (\theta^2 \sigma_{\bm{w}}^2 + \rho^2)\\ \frac{2}{3} (\theta u_i + \rho)^8 (\theta^2 \sigma_{\bm{w}}^2 + \rho^2)\\
-\frac{4}{27} (\theta u_i + \rho)^{10} (\theta^2 \sigma_{\bm{w}}^2 + \rho^2)\\
\frac{1}{81}(\theta u_i + \rho)^{12} (\theta^2 \sigma_{\bm{w}}^2 + \rho^2)
\end{bmatrix},
\end{eqnarray*}
where the vector $\bm{\tilde{x}}=[x^2,x^4,x^6,x^8,x^{10},x^{12},x^{14}]^\top$; (a) involves a number of steps, the details of which are omitted here for simplicity. The approximation comes from $\tanh(\alpha) \approx \alpha - \frac{1}{3} \alpha^3$.

Next, we take the expectation over $x$ and vector $\bm{u}$, respectively. This gives us
\begin{equation*}
\mathbb{E}_x \mathbb{E}_{\bm{v}} B \approx \frac{1}{3} \sigma_{\bm{z}}^2 \theta^2 c^2 \sum_{i=1}^N \left[(\theta u_i + \rho)^2 + (\theta^2 \sigma_{\bm{w}}^2 + \rho^2) \right] + \psi(c^4),
\end{equation*}
and
\begin{equation*}
\mathbb{E}_x \mathbb{E}_{\bm{w}} B \approx \frac{2N}{3} \sigma_{\bm{z}}^2 \theta^2 c^2(\theta^2 \sigma_{\bm{w}}^2 + \rho^2) + \psi(c^4).
\end{equation*}

The derivation of $C$, again, is very complex, but it can be shown that all the entries of $\mathbb{E}_x \mathbb{E}_{\bm{w}} C$ are $\psi(c^4)$. Therefore, we can simply write $\mathbb{E}_x \mathbb{E}_{\bm{w}} C = \psi(c^4)$.

Finally, we have
\begin{eqnarray*}
&&\hspace{-0.5cm} \bar{\mathcal{D}} = \mathbb{E}_x \mathbb{E}_{\bm{w}} \mathbb{E}_{\bm{z}} \mathcal{D} = \mathbb{E}_x \mathbb{E}_{\bm{w}}(A + B + C) \\
&&\hspace{-0.5cm} \approx \frac{N}{3}(\theta^2 \sigma_{\bm{w}}^2 \!+\! \rho^2)^2 c^2 \!+\! N(N\!-\!1)\Big[ \frac{1}{3} \rho^4 c^2 \!+\! \big(\theta^2 \rho^2 \sigma_{\bm{w}}^2 \!+\! \frac{\rho^4}{3}  \big)^2\Big]\\
&&\hspace{-0.5cm}  - \frac{2N}{3} \theta^2 c^2 \sigma_{\bm{w}}^4 \!+\! \frac{N}{3}\sigma_{\bm{w}}^4 c^2 \!+\! \frac{2N}{3} c^2\sigma_{\bm{z}}^2 \theta^2 (\theta^2 \sigma_{\bm{w}}^2 \!+\! \rho^2) + \psi(c^4).
\end{eqnarray*}
After some manipulations, we arrive at \eqref{eq:exp2_D} in Theorem \ref{thm:exp2}.

When $c<1$ and $\psi(c^4)$ is negligible, the optimal $\theta$ and $\rho$ that minimize $\bar{\mathcal{D}}$ can be found by setting the derivative of $\bar{\mathcal{D}}$ w.r.t. $\theta$ and $\rho$ to zero, yielding
\begin{eqnarray*}
&&\hspace{-0.6cm} \frac{\partial \bar{\mathcal{D}}}{\partial \theta} \!=\! \frac{N}{3} c^2 \sigma_{\bm{w}}^4   4\theta^3 \!-\! \frac{2N}{3} c^2 \sigma_{\bm{w}}^4 2\theta \!+\! \frac{2N}{3} c^2 (\sigma_{\bm{w}}^2+\sigma_{\bm{z}}^2) 2\theta \rho^2 + \\
&&\hspace{-0.6cm} \frac{4N(N\!-\!1)}{3}\sigma_{\bm{w}}^2 \theta \rho^6 \!+\! \frac{8N}{3}c^2 \sigma_{\bm{w}}^2\sigma_{\bm{z}}^2 \theta^3 \!+\! N(N\!-\!1) \sigma_{\bm{w}}^4   4\theta^3 \rho^4 \!=\! 0, \\
&&\hspace{-0.5cm} \frac{\partial \bar{\mathcal{D}}}{\partial \rho} \!=\! \frac{4N}{3} c^2 (\sigma_{\bm{w}}^2  \!+\! \sigma_{\bm{z}}^2)\theta^2 \rho  \!+\! 4N(N\!\!-\!\!1) \sigma_{\bm{w}}^2 \theta^2 \rho^5 \!+\! 4N(N\!\!-\!\!1)
\\
&&\hspace{-0.5cm} \cdot \sigma_{\bm{w}}^4 \theta^4 \rho^3  \!+\! \frac{N^2}{3}c^2   4\rho^3 \!+\! \frac{N(N-1)}{9}   8\rho^7 \triangleq \rho \xi(\rho) \!=\! 0.
\end{eqnarray*}

To satisfy ${\partial \bar{\mathcal{D}}}/{\partial \rho}=0$, either $\rho = 0$ or $\xi(\rho)$ should be $0$. It is easy to verify that
\begin{equation*}
\xi(\rho) \geq \frac{4N}{3} c^2(\sigma_{\bm{w}}^2+\sigma_{\bm{z}}^2)\theta^2 + \frac{4N}{3} c^2 \rho^2 > 0.
\end{equation*}

Thus, the optimal $\rho^* = 0$. Substituting $\rho = 0$ into ${\partial \bar{\mathcal{D}}}/{\partial \theta}=0$ gives us
\begin{equation*}
\theta(c^2\sigma_{\bm{w}}^4\theta^2 + 2c^2\sigma_{\bm{w}}^2\sigma_{\bm{z}}^2\theta^2 - c^2 \sigma_{\bm{w}}^4) = 0.
\end{equation*}

As dictated by \eqref{eq:MMSEpb}, an additional constraint of $\theta$ is $\theta \in \left[\frac{\sigma_{\bm{w}}^2}{\sigma_{\bm{w}}^2+\sigma_{\bm{z}}^2},\infty \right)$. Thus, for general $c \neq 0$, we have
\begin{equation*}
\sigma_{\bm{w}}^2 \theta^2 + 2\sigma_{\bm{z}}^2 \theta^2 - \sigma_{\bm{w}}^2 = 0
\end{equation*}
and the optimal $\theta^*=\sqrt{\frac{\sigma_{\bm{w}}^2}{\sigma_{\bm{w}}^2+2\sigma_{\bm{z}}^2}}$. It can be verified that $\theta=\sqrt{\frac{\sigma_{\bm{w}}^2}{\sigma_{\bm{w}}^2+2\sigma_{\bm{z}}^2}}$ falls into the region $\left[\frac{\sigma_{\bm{w}}^2}{\sigma_{\bm{w}}^2+\sigma_{\bm{z}}^2},\infty \right)$ because
\begin{equation*}
(\theta^*)^2 - \left(\frac{\sigma_{\bm{w}}^2}{\sigma_{\bm{w}}^2+\sigma_{\bm{z}}^2} \right)^2 = \frac{\sigma_{\bm{w}}^2\sigma_{\bm{z}}^4}{(\sigma_{\bm{w}}^2+2\sigma_{\bm{z}}^2)(\sigma_{\bm{w}}^2+\sigma_{\bm{z}}^2)^2} > 0.
\end{equation*}

For $\theta^*(\lambda^*) = \sqrt{\frac{\sigma_{\bm{w}}^2}{\sigma_{\bm{w}}^2+2\sigma_{\bm{z}}^2}}$ and $\rho^*(\lambda^*,\beta^*) = 0$, we have
\begin{equation*}
\lambda^* = \frac{1}{2\sigma_{\bm{w}}^2} + \frac{1}{2\sigma_{\bm{z}}^2} - \sqrt{\frac{1}{4\sigma_{\bm{z}}^4}+\frac{1}{2\sigma_{\bm{z}}^2\sigma_{\bm{w}}^2}},~~ \beta^* = 0,
\end{equation*}
and
\begin{eqnarray*}
\bar{\mathcal{D}}^{\text{MMSE}_{pb}}
\hspace{-0.2cm}& \approx &\hspace{-0.2cm} \frac{N}{3}\left(\frac{\sigma_{\bm{w}}^4}{\sigma_{\bm{w}}^2+2\sigma_{\bm{z}}^2} \right)^2 c^2 - \frac{2N}{3} \frac{\sigma_{\bm{w}}^4}{\sigma_{\bm{w}}^2+2\sigma_{\bm{z}}^2}c^2\sigma_{\bm{w}}^2+ \\
&&\hspace{-0.2cm} \frac{N}{3}\sigma_{\bm{w}}^4 c^2 + \sigma_{\bm{z}}^2 \frac{1}{\sigma_{\bm{w}}^2}\left(\frac{\sigma_{\bm{w}}^4}{\sigma_{\bm{w}}^2+2\sigma_{\bm{z}}^2} \right)^2 \left(\frac{2N}{3}c^2\right) \\
\hspace{-0.2cm}& \approx &\hspace{-0.2cm} \frac{2Nc^2\sigma_{\bm{z}}^2\sigma_{\bm{w}}^4}{3(\sigma_{\bm{w}}^2+2\sigma_{\bm{z}}^2)}.
\end{eqnarray*}

Further, combing ${\bar{\mathcal{D}}}^\text{ML} \approx \frac{2Nc^2}{3}\sigma_{\bm{z}}^2\sigma_{\bm{w}}^2$, we have
\begin{eqnarray}
\frac{\bar{\mathcal{D}}^\text{ML}-\bar{\mathcal{D}}^{\text{MMSE}_{pb}}}{\bar{\mathcal{D}}^\text{ML}} \approx \frac{2\sigma_{\bm{z}}^2}{\sigma_{\bm{w}}^2+2\sigma_{\bm{z}}^2}.
\end{eqnarray}

\bibliographystyle{IEEEtran}
\bibliography{References}

\end{document}